\def\year{2022}\relax
\documentclass[letterpaper]{article} % DO NOT CHANGE THIS
\usepackage{aaai22}  % DO NOT CHANGE THIS
\usepackage{times}  % DO NOT CHANGE THIS
\usepackage{helvet}  % DO NOT CHANGE THIS
\usepackage{courier}  % DO NOT CHANGE THIS
\usepackage[hyphens]{url}  % DO NOT CHANGE THIS
\usepackage{graphicx} % DO NOT CHANGE THIS
\usepackage{natbib}  % DO NOT CHANGE THIS AND DO NOT ADD ANY OPTIONS TO IT
\usepackage{caption} % DO NOT CHANGE THIS AND DO NOT ADD ANY OPTIONS TO IT
\DeclareCaptionStyle{ruled}{labelfont=normalfont,labelsep=colon,strut=off} % DO NOT CHANGE THIS
\usepackage{algorithm}
\usepackage{algorithmic}

\usepackage{graphicx}
\usepackage{subfigure}
\usepackage{booktabs}
\usepackage{amsmath}
\usepackage{amssymb}
\usepackage{bbm}
\usepackage{stfloats}
\usepackage[dvipsnames]{xcolor}
\usepackage{enumitem}
\usepackage{tikz}
\usepackage{multirow}
\usepackage{sidecap}
\usepackage{placeins}
\usepackage{hyperref}
\usepackage{parskip}

\usepackage{newfloat}
\usepackage{listings}
\floatstyle{ruled}
\newfloat{listing}{tb}{lst}{}
\floatname{listing}{Listing}
\newcommand{\attr}{$\mathrm{\%Attr}$}
\newcommand{\er}{$\mathrm{\%ER}$}
\newcommand{\sel}{$\mathrm{\%Sel}$}
\newcommand{\gb}[1]{\textcolor{Green}{\textbf{#1}}}
\newcommand{\ri}[1]{\textcolor{Red}{\textit{#1}}}
\renewcommand{\u}[1]{\underline{#1}}

\let\vec\mathbf
\allowdisplaybreaks

\usepackage{graphbox}

\graphicspath{{figures/}}

\title{Do Feature Attribution Methods Correctly Attribute Features?}
\author{
    Yilun Zhou$^1$, Serena Booth$^1$, Marco Tulio Ribeiro$^2$, Julie Shah$^1$
}
\affiliations{
    $^1$MIT CSAIL, $^2$Microsoft Research\\
    $^1$\{yilun, serenabooth, julie\_a\_shah\}@csail.mit.edu, $^2$marcotcr@microsoft.com
}

\begin{document}

\maketitle

\begin{abstract}
Feature attribution methods are popular in interpretable machine learning. These methods compute the attribution of each input feature to represent its importance, but there is no consensus on the definition of ``attribution'', leading to many competing methods with little systematic evaluation, complicated in particular by the lack of ground truth attribution. To address this, we propose a dataset modification procedure to induce such ground truth. Using this procedure, we evaluate three common methods: saliency maps, rationales, and attentions. We identify several deficiencies and add new perspectives to the growing body of evidence questioning the correctness and reliability of these methods applied on datasets in the wild. We further discuss possible avenues for remedy and recommend new attribution methods to be tested against ground truth before deployment. The code is available at \url{https://github.com/YilunZhou/feature-attribution-evaluation}. 
\end{abstract}

\vspace{-0.1in}
\section{Introduction}
\label{sec:introduction}
Consider the task of training a neural network to detect cancers from X-ray images, wherein the data come from two sources: a general hospital and a specialized cancer center. As can be expected, images from the cancer center contain many more cancer cases, but imagine the cancer center adds a small timestamp watermark to the top-left corner of its images. Since the timestamp is a strongly correlated with cancer presence, a model may learn to use it for prediction.

It is important to ensure the deployed model makes predictions based on genuine medical signals rather than image artifacts like watermarks. If these artifacts are known \textit{a priori}, we can evaluate the model on counterfactual pairs---images with and without them---and compute prediction difference to assess their impact. However, for almost all datasets, we cannot realistically anticipate every possible artifact. As such, feature attribution methods like saliency maps \citep{simonyan2013deep}
are used to identify regions which are important for prediction, which humans then inspect for evidence of any artifacts. This train-and-interpret pipeline has been widely adopted in data-driven medical diagnosis \citep{shen2019deep, mostavi2020convolutional, si2021fully} and many other applications.

\begin{figure}[!t]
    \centering
    \includegraphics[width=1\columnwidth]{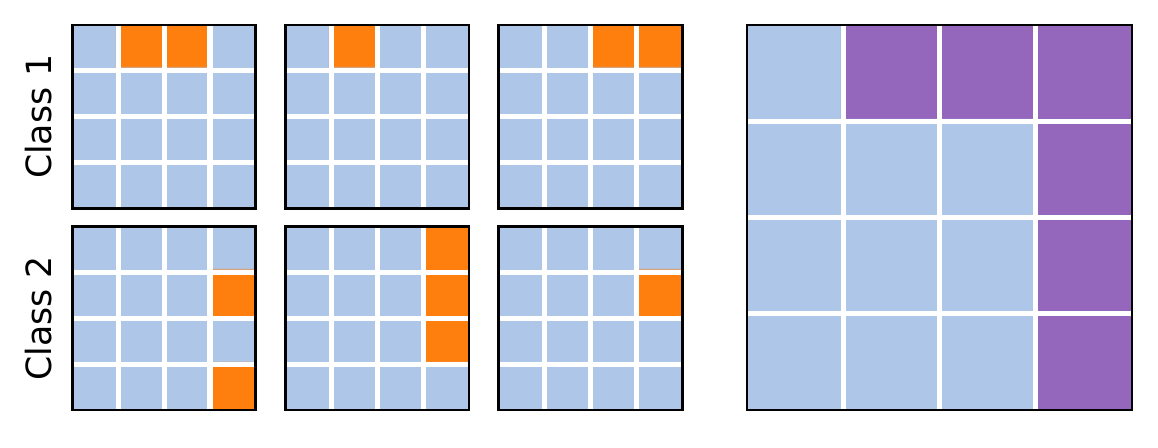}
    \caption{The intuition behind our feature attribution ground truth: if we know that for every input, only specific features (\textcolor[RGB]{255,127,14}{orange}) are informative to the label, then across the dataset, a high-performing model has to focus on them and not get ``distracted'' by other irrelevant features. Thus, feature attributions should highlight the union \textit{union} of these features (\textcolor[RGB]{148,103,189}{purple}), and any attribution outside this area is misleading. 
    }
    \label{fig:overall-diagram}
\end{figure}

Crucially, this procedure assumes that the attribution methods works correctly and does not miss influential features. Is this truly the case? Direct evaluation on natural datasets is impossible as the very spurious correlations we want attribution methods to find are, by definition, unknown. Many evaluations try to sidestep this problem with proxy metrics \citep{samek2017evaluating, hooker2018benchmark, bastings2019interpretable}, but they are limited in various ways, notably by a lack of ground truth, as discussed in Sec.~\ref{sec:related-work-evaluation}.

Instead, we propose evaluating these attribution methods on \textit{semi-natural} datasets: natural datasets systematically modified to introduce ground truth information for attributions. This modification (Fig.~\ref{fig:overall-diagram}) ensures that \textit{any} classifier with sufficiently high performance has to rely, sometimes solely, on the manipulations. We then present desiderata, or necessary conditions, for correct attribution values; for example, features known not to affect the model's decision should not receive attribution. The high-level idea is domain-general, and we instantiate it on image and text data to evaluate saliency maps, rationale models and attention mechanisms used to explain common deep learning architectures. We identify several failure modes of these methods, discuss potential reasons and recommend directions to fix them. Last, we advocate for testing new attribution methods against ground truth to validate their attributions before deployment.

\section{Related Work}
\label{sec:related-work}

\subsection{Feature Attribution Methods}
\label{app:related-work-methods}

\label{sec:related-work-saliency}
\label{sec:related-work-rationale}
\label{sec:related-work-attention}

Feature attribution methods assign attribution scores to input features, the absolute value of which informally represents their importance to the model prediction or performance. 

\noindent\textbf{Saliency maps} explain an image $I$ by producing $S$ of the same size, where $S_{h, w}$ indicates the contribution of pixel $I_{h, w}$. In various works, the notion of contribution has been defined as sensitivity \citep{simonyan2013deep}, relevance \citep{bach2015pixel}, local influence \citep{ribeiro2016should}, Shapley values \citep{lundberg2017unified}, or filter activations \citep{selvaraju2017grad}. 

\noindent\textbf{Attention mechanisms} \citep{bahdanau2014neural} were originally proposed to better retain sequential information. Recently they have been used as attribution values, but their their validity is under debate with different and inconsistent criteria being proposed \citep{jain2019attention, wiegreffe2019attention, pruthi2020learning}. 

\noindent\textbf{Rationale models} \citep{lei2016rationalizing, bastings2019interpretable, jain2020learning} are inherently interpretable models for text classification with a two-stage pipeline: a selector extracts a rationale (i.e. input words), and a classifier makes a prediction based on it. The selected rationales are often regularized to be succinct and continuous. 

\subsection{Evaluation of Feature Attributions}
\label{sec:related-work-evaluation}
At their core, feature attribution methods describe mathematical properties of the model's decision function. For example, gradient describes sensitivity with respect to infinitesimal input perturbation, and SHAP describes a notion of values in a multi-player game with features as players. We associate these mathematical properties with high-level interpretations such as ``feature importance'', and it is this association that requires justification. 

A popular way is to assess alignment with human judgment, but models and humans can reach the same prediction while using distinct reasoning mechanisms (e.g. medical signals used by doctors and watermarks used by the model). For example, SmoothGrad \citep{smilkov2017smoothgrad} is proposed as an improvement to the original Gradient \citep{simonyan2013deep} since it gives less noisy and more legible saliency maps, but it is not clear whether saliency maps \emph{should} be smooth. \citet{bastings2019interpretable} evaluated their rationale model by assessing its agreement with human rationale annotation, but a model may achieve high accuracy with subtle but strongly correlated textual features such as grammatical idiosyncrasy. \citet{covert2020understanding} compared the feature attribution of a cancer prediction model to scientific knowledge, yet a well-performing model may rely on other signals. In general, positive results from alignment evaluation only support plausibility \citep{jacovi2020towards}, not faithfulness. 

Another common approach successively removes features with the highest attribution values and evaluates certain metrics. One metric is prediction change \citep[e.g.][]{samek2017evaluating, arras2019evaluating, ismail2020benchmarking}, but it fails to account for nonlinear interactions: for an OR function of two active inputs, the evaluation will (incorrectly) deem whichever feature removed first to be useless as its removal does not affect the prediction. Another metric is model retraining performance \citep{hooker2018benchmark}, which may fail when different features lead to the same accuracy---as is often possible~\cite{d2020underspecification}. For example, a model might achieve some accuracy by using only feature $x_1$. If a retrained model using only $x_2$ achieves the same accuracy, the evaluation framework would (falsely) reject the ground truth attribution of $x_1$ due to the same re-training accuracy. 

Most similar to our proposal are works that also construct semi-natural datasets with explicitly defined ground truth explanations \citep{yang2019benchmarking, adebayo2020debugging}. \citet{adebayo2020debugging} used a perfect background correlation for a dog-vs-bird dataset, found that the model achieves high accuracy on background alone, and claimed that the correct attribution should focus solely on the background. However, we verified that a model trained on their dataset can achieve high accuracy \textit{simultaneously} on foreground alone, background alone, and both combined, invalidating their ground truth claim. Similarly, \citet{yang2019benchmarking} argue that for \textit{background} classification, a label-correlated foreground should receive high attribution value, but a model could always rely solely on background with perfect label correlation. We avoid such pitfalls via label reassignment (Sec.~\ref{sec:reassignment}), so that the model \textit{must} use target features for high accuracy. Furthermore, a more subtle failure mode, in which the model can (rightfully) use the absence of information for a prediction, is avoided by our joint effective region formulation, discussed in the Remark at the end of Sec.~\ref{sec:modification}. 

Finally, \citet{adebayo2018sanity} proposed sanity checks for saliency maps by assessing their change under weight or label randomization. We establish complementary criteria for explanations by instead focusing on model-agnostic dataset-side modifications, and identify additional failure cases.

\section{Desiderata for Attribution Values}
\label{sec:desiderata}
What should the attribution values be? Although the precise values may be axiomatic, certain properties are \textit{de facto} requirements if we want people to understand how a model makes a decision, verify that its reasoning process is sound, and possibly inform options for correction if it is not (c.f. the opening example in Sec.~\ref{sec:introduction}). For example, while LIME and SHAP define attribution differently, both would produce undeniably bad explanations if they highlight features completely ignored by the model. 

We study two types of features: those of fundamental importance to the model, denoted by $F_C$, and those non-informative to the label, denoted by $F_N$. A first requirement is that explanations should not miss important features, $F_C$. Unfortunately, identifying all such features is not easy. For example, while the model could potentially use the timestamp on some X-ray images for cancer prediction, it could instead exclusively rely on genuine medical features (as done by human doctors), and attributions should only highlight the timestamp in the former case. This difficulty motivates our dataset modification procedure detailed in the next section. In brief, we can modify the dataset such that any model using only medical features could not achieve a high accuracy (due to introduced label noise), thus establishing the ground truth usage of the timestamp for any model with high accuracy. We can then evaluate how well the attribution method identifies the contribution of the timestamp by the attribution percentage $\mathrm{Attr}\%$ of the timestamp pixels, with
$\mathrm{Attr\%}(F) \doteq \left(\sum_{i\in F}{|s_i|}\right) / \left(\sum_{i=1}^D{|s_i|}\right)$, where $D$ is the total number of features and $s_i$ is the attribution value assigned to the $i$-th feature. Since $F_C$ contains all features used by the model, we should expect $\mathrm{Attr\%}(F_C)\approx 1$. 

Conversely, we can introduce non-informative features $F_N$ independent from the label---for example, a white border added to randomly selected images. While the model prediction could depend on it (e.g. more positive for those with the border), methods that study features contributing to the \textit{performance} should not highlight $F_N$. In addition, any reliance on $F_N$ is detrimental to performance, and as performance increases, a good prediction is less ``distracted'' by $F_N$, which should correspondingly not get highlighted, or in other words $\mathrm{Attr}\%(F_N)$ should decrease to 0. 

In addition to continuous attributions on all features, another formulation selects $k$ features, with no distinction among them. This can either be derived by a top-$k$ post-processing to induce sparse explanations, or generated directly by some models, e.g. as rationales \citep{lei2016rationalizing}. For the first case, a hyper-parameter $k$ needs to be chosen. A small value risks missing important features while a large value may include unnecessary features that obfuscate true model reasoning. For the second case, $k$ is typically chosen automatically by the model, e.g. the rationale selector. In both cases, ensuring that $F_C$ is highlighted (i.e. $\mathrm{Attr\%} \allowbreak = \allowbreak 1$) is easily ``hackable'' by just selecting all features. As such, we instead use two information-retrieval metrics, precision and recall, defined as $\mathrm{Pr}(F) = |F \cap F_C| / |F|$, and $\mathrm{Re}(F) = |F \cap F_C| / |F_C|$ for evaluating these attributions, where $F$ is the $k$ selected features.

\section{Dataset Modification with Ground Truth}
\label{sec:modification}

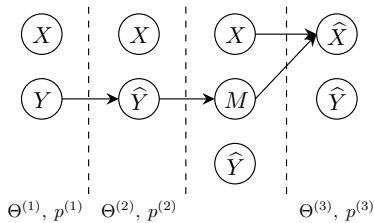
\begin{figure}[!htb]
    \centering
    \resizebox{0.6\columnwidth}{!}{\tikzset{every picture/.style={line width=0.75pt}} %set default line width to 0.75pt        

\begin{tikzpicture}[x=0.75pt,y=0.75pt,yscale=-1,xscale=1]
%uncomment if require: \path (0,494); %set diagram left start at 0, and has height of 494

%Shape: Circle [id:dp5113925153554704] 
\draw   (71,94.5) .. controls (71,84.84) and (78.84,77) .. (88.5,77) .. controls (98.16,77) and (106,84.84) .. (106,94.5) .. controls (106,104.16) and (98.16,112) .. (88.5,112) .. controls (78.84,112) and (71,104.16) .. (71,94.5) -- cycle ;

%Shape: Circle [id:dp37657066042032494] 
\draw   (71,150.5) .. controls (71,140.84) and (78.84,133) .. (88.5,133) .. controls (98.16,133) and (106,140.84) .. (106,150.5) .. controls (106,160.16) and (98.16,168) .. (88.5,168) .. controls (78.84,168) and (71,160.16) .. (71,150.5) -- cycle ;
%Shape: Circle [id:dp34344459922238935] 
\draw   (155,94.5) .. controls (155,84.84) and (162.84,77) .. (172.5,77) .. controls (182.16,77) and (190,84.84) .. (190,94.5) .. controls (190,104.16) and (182.16,112) .. (172.5,112) .. controls (162.84,112) and (155,104.16) .. (155,94.5) -- cycle ;

%Shape: Circle [id:dp9630558506209146] 
\draw   (155,150.5) .. controls (155,140.84) and (162.84,133) .. (172.5,133) .. controls (182.16,133) and (190,140.84) .. (190,150.5) .. controls (190,160.16) and (182.16,168) .. (172.5,168) .. controls (162.84,168) and (155,160.16) .. (155,150.5) -- cycle ;
%Straight Lines [id:da5874273922159872] 
\draw    (106,150.5) -- (152,150.5) ;
\draw [shift={(155,150.5)}, rotate = 180] [fill={rgb, 255:red, 0; green, 0; blue, 0 }  ][line width=0.08]  [draw opacity=0] (10.72,-5.15) -- (0,0) -- (10.72,5.15) -- (7.12,0) -- cycle    ;
%Shape: Circle [id:dp5401433684545605] 
\draw   (238,94.5) .. controls (238,84.84) and (245.84,77) .. (255.5,77) .. controls (265.16,77) and (273,84.84) .. (273,94.5) .. controls (273,104.16) and (265.16,112) .. (255.5,112) .. controls (245.84,112) and (238,104.16) .. (238,94.5) -- cycle ;

%Shape: Circle [id:dp08354080352674909] 
\draw   (238,206.5) .. controls (238,196.84) and (245.84,189) .. (255.5,189) .. controls (265.16,189) and (273,196.84) .. (273,206.5) .. controls (273,216.16) and (265.16,224) .. (255.5,224) .. controls (245.84,224) and (238,216.16) .. (238,206.5) -- cycle ;
%Shape: Circle [id:dp00026554622237129344] 
\draw   (238,150.5) .. controls (238,140.84) and (245.84,133) .. (255.5,133) .. controls (265.16,133) and (273,140.84) .. (273,150.5) .. controls (273,160.16) and (265.16,168) .. (255.5,168) .. controls (245.84,168) and (238,160.16) .. (238,150.5) -- cycle ;
%Straight Lines [id:da6578966509899598] 
\draw    (190,150.5) -- (235,150.5) ;
\draw [shift={(238,150.5)}, rotate = 180] [fill={rgb, 255:red, 0; green, 0; blue, 0 }  ][line width=0.08]  [draw opacity=0] (10.72,-5.15) -- (0,0) -- (10.72,5.15) -- (7.12,0) -- cycle    ;
%Shape: Circle [id:dp5926862825283326] 
\draw   (326,94.5) .. controls (326,84.84) and (333.84,77) .. (343.5,77) .. controls (353.16,77) and (361,84.84) .. (361,94.5) .. controls (361,104.16) and (353.16,112) .. (343.5,112) .. controls (333.84,112) and (326,104.16) .. (326,94.5) -- cycle ;
%Shape: Circle [id:dp06464059609796213] 
\draw   (326,150.5) .. controls (326,140.84) and (333.84,133) .. (343.5,133) .. controls (353.16,133) and (361,140.84) .. (361,150.5) .. controls (361,160.16) and (353.16,168) .. (343.5,168) .. controls (333.84,168) and (326,160.16) .. (326,150.5) -- cycle ;
%Straight Lines [id:da2612584926033552] 
\draw    (273,150.5) -- (323.94,96.68) ;
\draw [shift={(326,94.5)}, rotate = 493.42] [fill={rgb, 255:red, 0; green, 0; blue, 0 }  ][line width=0.08]  [draw opacity=0] (10.72,-5.15) -- (0,0) -- (10.72,5.15) -- (7.12,0) -- cycle    ;
%Straight Lines [id:da954637997777944] 
\draw    (273,94.5) -- (323,94.5) ;
\draw [shift={(326,94.5)}, rotate = 180] [fill={rgb, 255:red, 0; green, 0; blue, 0 }  ][line width=0.08]  [draw opacity=0] (10.72,-5.15) -- (0,0) -- (10.72,5.15) -- (7.12,0) -- cycle    ;
%Straight Lines [id:da25574613187452666] 
\draw  [dash pattern={on 4.5pt off 4.5pt}]  (129,70.8) -- (129,232.8) ;
%Straight Lines [id:da25007084692594694] 
\draw  [dash pattern={on 4.5pt off 4.5pt}]  (213,70.8) -- (213,232.8) ;
%Straight Lines [id:da8714282385622332] 
\draw  [dash pattern={on 4.5pt off 4.5pt}]  (300,70.8) -- (300,232.8) ;

% Text Node
\draw (78,86.4) node [scale=1.5][anchor=north west][inner sep=0.75pt]    {$X$};
% Text Node
\draw (79,142.4) node [scale=1.5][anchor=north west][inner sep=0.75pt]    {$Y$};
% Text Node
\draw (163,138.4) node [scale=1.5][anchor=north west][inner sep=0.75pt]    {$\widehat{Y}$};
% Text Node
\draw (162,86.4) node [scale=1.5][anchor=north west][inner sep=0.75pt]    {$X$};
% Text Node
\draw (246,194.4) node [scale=1.5][anchor=north west][inner sep=0.75pt]    {$\widehat{Y}$};
% Text Node
\draw (245,86.4) node [scale=1.5][anchor=north west][inner sep=0.75pt]    {$X$};
% Text Node
\draw (244,142.4) node [scale=1.5][anchor=north west][inner sep=0.75pt]    {$M$};
% Text Node
\draw (334,138.4) node [scale=1.5][anchor=north west][inner sep=0.75pt]    {$\widehat{Y}$};
% Text Node
\draw (333,82.4) node [scale=1.5][anchor=north west][inner sep=0.75pt]    {$\widehat{X}$};
% Text Node
\draw (57,236.2) node [scale=1.2][anchor=north west][inner sep=0.75pt]    {$\Theta ^{( 1)} ,\ p^{( 1)}$};
% Text Node
\draw (138,236.2) node [scale=1.2][anchor=north west][inner sep=0.75pt]    {$\Theta ^{( 2)} ,\ p^{( 2)}$};
% Text Node
% \draw (224,236.2) node [scale=1.2][anchor=north west][inner sep=0.75pt]    {$\Theta ^{( 3)} ,\ p^{( 3)}$};
% Text Node
\draw (309,236.2) node [scale=1.2][anchor=north west][inner sep=0.75pt]    {$\Theta ^{( 3)} ,\ p^{( 3)}$};

\end{tikzpicture}}
    \caption{The graphical model for our dataset modification. }
    \label{fig:dm-graphical-model}
\end{figure}

We now present the dataset modification procedure that lets us quantify the influence of certain features to the model. We use a running example of adding a watermark pattern to a watermark-free X-ray cancer dataset, such that the newly added watermark is guaranteed to affect the model decision. 

Let $\mathcal X$ and $\mathcal Y \doteq \{1, ..., K\}$ be input and output space for $K$-class classification. Fig.~\ref{fig:dm-graphical-model} shows two modification steps: from an original data instance (1st column), label reassignment reduces the predictive power of existing signals (2nd column) and input manipulation introduces new predictive features (3rd and 4th columns). 

\noindent \textbf{Label Reassignment\quad}
\label{sec:reassignment}
Our goal is to ensure that the model has to rely on certain introduced features (e.g. a watermark) to achieve a high performance. However, the model could in theory use any of the existing features (e.g. medical features) to achieve high accuracy, and thus disregard the new feature, even if it is perfectly correlated with the label. To guarantee the model's usage of new features, we need to weaken the correlation between the original features and the labels. 

We first consider label reassignment for binary classification, which is used in all experiments. During reassignment, the label is preserved with probability $r$ and flipped otherwise, so the accuracy without relying on the manipulation is at most $p^*=\max(r, 1-r)$. For the special case of $r=0.5$, no features are informative to the label, and the performance is random in expectation. After label reassignment, a data point $(x,y)$ becomes $(x, \widehat y)$.

More generally, the $K$-class setting is modeled by a reassignment matrix $R\in \mathbb R^{K\times K}$. According to this matrix, the label reassignment process assigns a new label $\widehat y$ based on the original label $y$ with probability $R_{y, \widehat y}$. The expected accuracy $p^{(2)}$ of \textit{any} classifier is bounded by $p^* = \max_{i, j}R_{i, j}$. 

\noindent \textbf{Input Manipulation\quad}
Next, we apply manipulations on the input $x$ according to its reassigned label $\widehat y$. We consider a set of $L$ input manipulations, $\mathcal M=\{m_1, ...,m_L\}$, and a manipulation function $q: \mathcal M\times \mathcal X\rightarrow \widehat{\mathcal X}$ such that $q(m_l, x)=\widehat x$ applies the manipulation on the input and returns the manipulated output $\widehat x$. 
$\mathcal M$ can include the blank manipulation $m_\varnothing$ that leaves the input unchanged.

To facilitate feature attribution evaluation, we require the manipulation to be \textit{local}, in that $q(m, x)$ affects only a part of the input $x$. Formally, 
we define the \textit{effective region} (ER) of $m_l$ on $x$ as the set of input features modified by $m_l$, denoted as $\phi_l(x) \doteq \left\{i: \left[q(m_l, x)\right]_{i} \neq x_{i}\right\},$ where subscript $i$ indexes over individual features (e.g. pixels). The blank manipulation has empty ER, $\phi_{\varnothing} = \varnothing$.

For $(x, \widehat y)\sim \mathbb P_{X, \widehat Y}$, we choose a manipulation $m_l$ from $\mathcal M$ according to $\widehat y$ and modify the input as $\widehat x = q(m_l, x)$. The label-dependent choice can be deterministic or stochastic. We denote the new data distribution as $\mathbb P_{\widehat X, \widehat Y}$. With appropriate choice of manipulation, $\mathbb P_{\widehat X, \widehat Y}$ can satisfy $\widehat{p}^{\,*}\doteq \sup_{\widehat x, \widehat y}\allowbreak \mathbb P_{\widehat Y|\widehat X}(\widehat y|\widehat x) > p^*$. For example, $\widehat p^{\,*} = 1$ is achievable when a watermark is applied exclusively to the positive class.

\emph{Whenever a model trained on $(\widehat{\mathcal X}, \widehat{\mathcal Y})$ achieves expected accuracy $p^{(3)} > p^*$, it is guaranteed to rely on the knowledge of manipulation, which is solely confined within the \textit{joint effective region} $\phi_\cup(x) \doteq \cup_l \,\phi_l(x)$.} This gives us a straightforward, quantitative check for feature attribution methods: they should recognize the contribution inside $\phi_\cup(x)$. For our example, since only the watermark is applied to one class, $\phi_\cup$ corresponds to the watermarked region. 

On finite test sets, a classifier can achieve an accuracy $p > p^*$ without using the manipulation, due to stochasticity in label reassignment. However, for test set size $N$, the probability of this classifier achieving of $p$ or higher, when the \textit{expected} accuracy is bounded by $p^*$, is at most $\sum_{n=\lfloor pN\rfloor}^N \mathrm{Binom}(n; N, p^*)$, which vanishes quickly with increasing $N$ and $p$. 

\noindent \textbf{Remark\quad} It is crucial to consider the \textit{joint} effective region over all manipulations for attribution values, since a model could use the \textit{absence} of manipulation as a legitimate basis for decision. For example, consider an image dataset, with each image having a watermark either on the top or bottom edge correlated with the positive or negative label respectively. A model could make negative predictions based on the \textit{absence} of a watermark on the top edge. In this case, the correct attribution to the top edge is within the joint ER but \textit{not} within the bottom watermark ER. Current evaluations \citep{yang2019benchmarking, adebayo2020debugging} often omit this possibility by using the ER of \textit{only} the manipulation applied to the target class rather than the union of all possible ERs for every class, potentially rejecting correct attributions. A more detailed explanation of how our proposed work differs from and improves upon those of \citet{yang2019benchmarking} and \citet{adebayo2020debugging} is provided in App.~\ref{app:rw-comparison}.

In next three sections, we experimentally compare attribution values of three types of models---saliency maps, attention mechanisms and rationale models---to those expected by the desiderata. Through the analysis, we identify their deficiencies and give recommendations for improvements.

\section{Evaluating Image Saliency Maps}

For these experiments, we simulate a common scenario where a model seemingly achieves ``superhuman'' performance on some hard image classification task, only for us to later find out that it exploits some image artifacts which are accidentally leaked in during the data collection process. We evaluate the extent to which several different saliency map attribution methods can identify such artifacts. 

\begin{figure}[!b]
\centering
\includegraphics[width=\columnwidth]{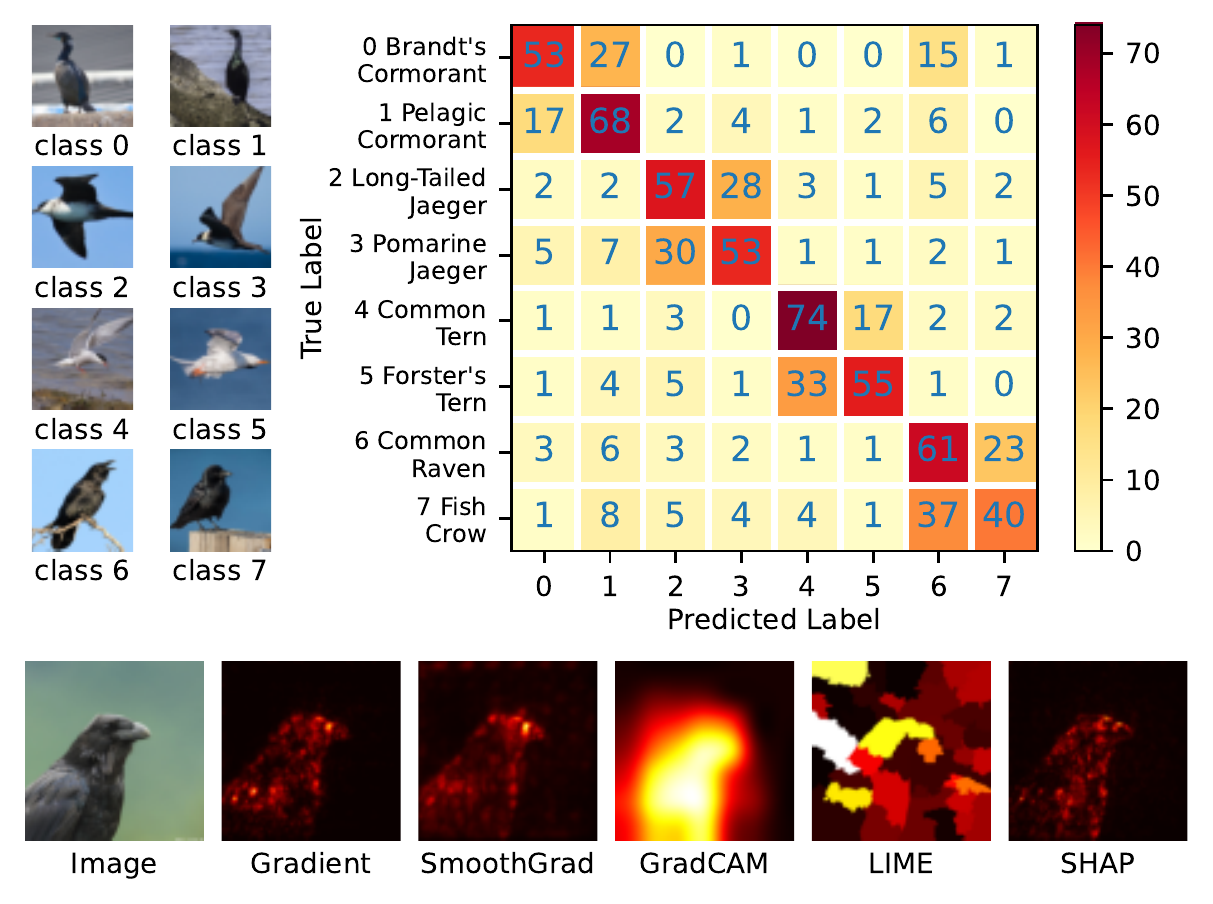}
\caption{Top: Dataset samples and test set confusion matrix of a ResNet-34 model. Bottom: Examples of five different saliency maps for a correct prediction (Fish Crow). }
\label{fig:flickr-bird-overview}
\end{figure}

\noindent \textbf{Model: }We used the ResNet-34 architecture \citep{he2016deep} for all experiments. The parameters are randomly initialized rather than pre-trained on ImageNet~\citep{deng2009imagenet}. 

\noindent \textbf{Dataset: }We curate our own dataset on bird species identification. First, we train a ResNet-34 model on CUB-200-2011 \citep{WahCUB_200_2011} and identify the top four most confusing class pairs. Then, we scrape Flickr for 1,200 new images per class, center-crop all images to $224\times224$ and mean-variance normalize using ImageNet statistics. Last, we split the 1,200 images per class into train/validation/test sets of 1000/100/100 images. Fig.~\ref{fig:flickr-bird-overview} presents sample images, the confusion matrix for a ResNet-34 model trained on this data, and example saliency maps for a correct prediction.

\begin{figure}[!htb]
    \centering
    \includegraphics[width=\columnwidth]{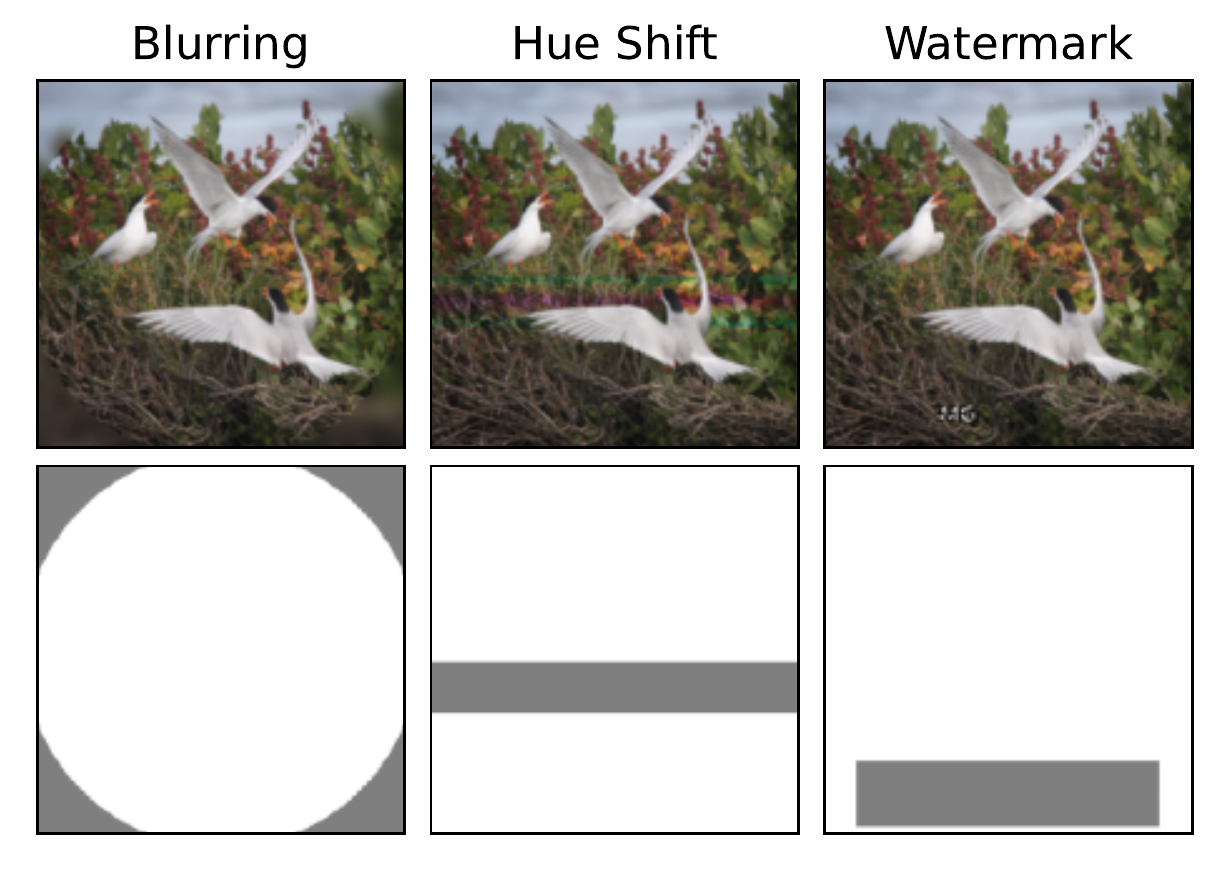}
    \caption{Three manipulations applied on the same image and their effective regions (gray). }
    \label{fig:img-manipulation}
\end{figure}

\noindent\textbf{Input Manipulations: }We define five image manipulations which represent artifacts that could be accidentally introduced in a dataset collection process: blurring, brightness change, hue shift, pixel noise, and watermark. Fig.~\ref{fig:img-manipulation} shows the effect of for three manipulations along with the effective regions. Other manipulation types and additional details are presented in in App.~\ref{app:image-correlation}. 

\noindent\textbf{Saliency Maps: }We evaluate 5 saliency map methods: Gradient \citep{simonyan2013deep}, SmoothGrad \citep{smilkov2017smoothgrad}, GradCAM \citep{selvaraju2017grad}, LIME \citep{ribeiro2016should}, and SHAP \citep{lundberg2017unified}, detailed in  App.~\ref{app:eval-saliency-map-methods}. 

\noindent\textbf{Experiments: } We set up binary classifications with pairs of easily confused species (e.g. common tern and Forester's tern) to simulate a hard task which is made easier through the presence of artifacts. Sec.~\ref{sec:saliency-attr-m} also uses pairs of visually distinct species (e.g. common tern and fish crow).

\noindent\textbf{Metric: }We study the attribution percentage assigned to the joint effective region $\mathrm{Attr}\%(\phi_\cup)$. We calculate \attr{} for images in the test set, and report the average separately for images of the two classes. 

\subsection{Attr\% by Attributions and Manipulations}
\label{sec:saliency-attr-ba}
\textbf{Question: }How well do saliency maps give attribution to the ground truth for (near-)perfect models? 

\textbf{Setup: }We train 100 models, each on a random pair of similar species and a random manipulation type. We reassign labels with $r=0.5$ (i.e. totally randomly), and apply the manipulation to images of the positive post-reassignment class, leaving the negative class images unchanged. 

\textbf{Expectation: }
With $r=0.5$, \emph{only} the manipulation is correlated with the label. A near-perfect performance thus indicates that the model relies almost exclusively on features inside $\phi_\cup$. Thus, we should expect $\mathrm{Attr}\%(F_{\phi_\cup})\approx 1.0$, regardless of the size of $\phi_\cup$. 

\begin{figure}[!t]
    \centering
    \includegraphics[width=\columnwidth]{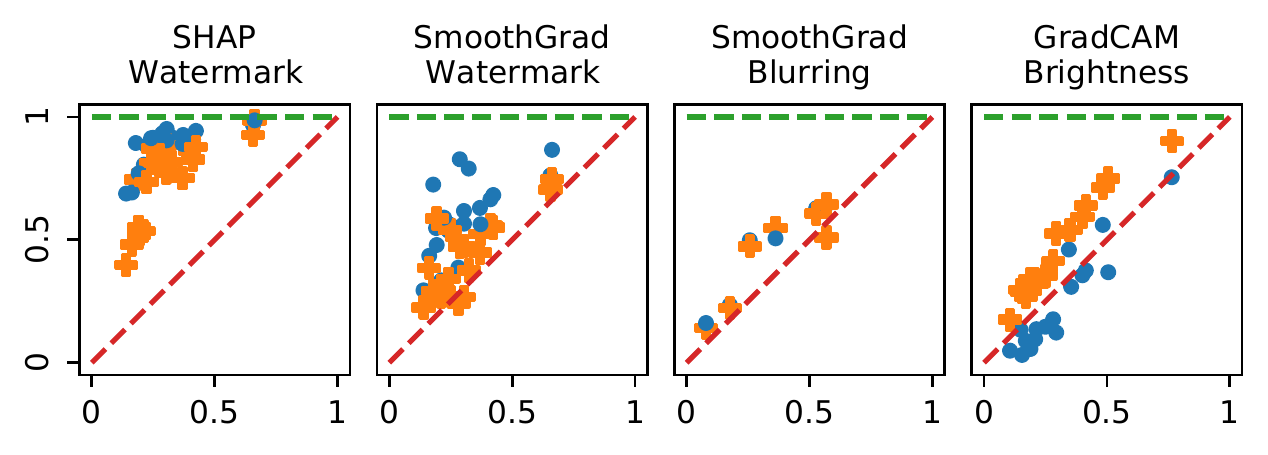}
    \caption{\attr{} ($y$-axis) vs. \er{} ($x$-axis), complete results in Fig.~\ref{fig:saliency-map-attr-ba-complete} of App.~\ref{app:eval-saliency-map-attr-ba}. \textcolor[RGB]{31,119,180}{Blue} circles and \textcolor[RGB]{255,127,14}{orange} crosses are for images with and without the manipulation. \textcolor[RGB]{44,160,44}{Green} horizontal line indicates the saliency map with $\mathrm{Attr\%}=1$ and \textcolor[RGB]{214,39,40}{red} diagonal line indicates a random saliency map. 
    }
    \label{fig:saliency-map-attr-ba}
\end{figure}

\textbf{Results: }
70\% of all runs achieve test accuracy of over 95\%\footnote{Note that since the model is not 100\% accurate, it could be ``distracted'' by features outside of $F_C$. However, such distraction is small, accounting for at most 5\% of errors, and it is much more important for users to understand that the over 95\% accuracy comes solely from $F_C$, and thus requiring $\mathrm{Attr\%}\approx 1$ is reasonable. }. We compute $\mathrm{Attr}\%(F_{\phi_\cup})$ for these models. Since \attr{} naturally depends on the size of $\phi_\cup$ (e.g. $\phi_\cup$ of the entire image implies $\mathrm{Attr}\% = 1$), we plot them against \er{}, defined as the size of $\phi_\cup$ as a fraction of image size. Fig.~\ref{fig:saliency-map-attr-ba} shows these two values for some methods and manipulations (complete results in
App.~\ref{app:eval-saliency-map-attr-ba}).

The models successfully learn all manipulation types, as demonstrated by the high test accuracy. However, none of the methods consistently scores \attr{} $\approx 1$. Further, not all manipulations are equally well detected by all saliency maps. While SHAP performs the best (\attr{} $ = 69\%$ at \er{} $= 40\%$ on average), it is still hard to trust ``in the wild'' since its efficacy strongly depends on manipulation type. 
The \emph{presence} of a watermark is often better detected than its \emph{absence}, likely because the model implicitly localizes objects (i.e. the watermark) \citep{bau2017network} and predicts a default negative class if it fails to do so. It is also easier for perturbation-based methods such as LIME to ``hide'' it when present than to ``construct'' it when absent. Thus, saliency maps may mislead people about the true reason for a negative prediction, and better methods to convey the absence of signals are needed. 

\subsection{Attribution vs. Test Accuracy}
\textbf{Question: }
How does \attr{} change as the model's test accuracy increases during training? 

\noindent\textbf{Setup: }
We use the the same setup as Sec.~\ref{sec:saliency-attr-ba}. 

\noindent\textbf{Expectation: }
As the test accuracy increases, the model must also increasingly rely on knowledge of manipulation. As a result, we should expect $\mathrm{Attr}\%(F_{\phi_\cup})$ to increase. 

\begin{figure}[!t]
    \centering
    \includegraphics[width=\columnwidth]{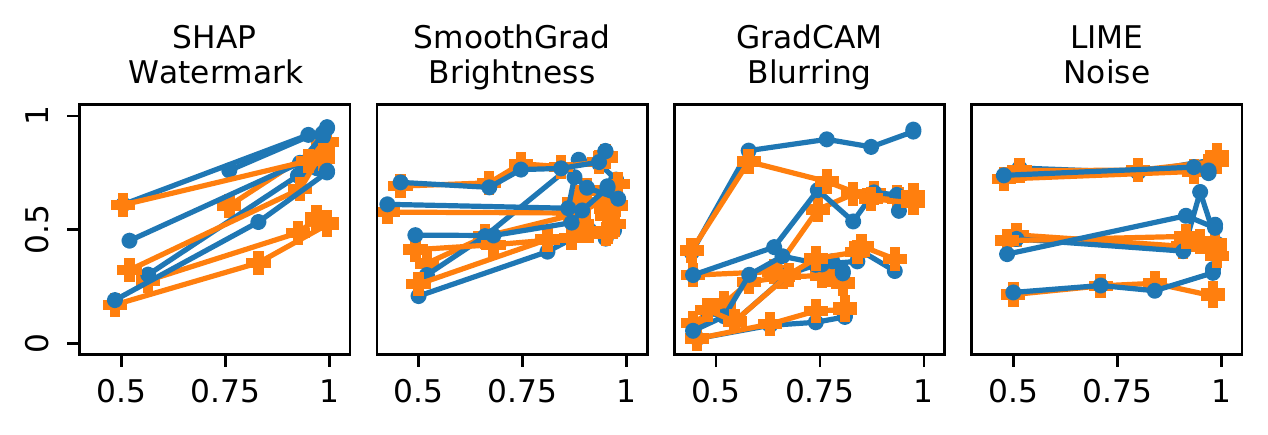}
    \caption{\attr{} vs. test accuracy, more in  App.~\ref{app:eval-saliency-map-test-acc}. }
    \label{fig:saliency-map-test-acc}
\end{figure}

\noindent\textbf{Results: }
For the training run of each model, we compute $\mathrm{Attr}\%(F_{\phi_\cup})$ for models during intermediate epochs with various test accuracy scores. Fig.~\ref{fig:saliency-map-test-acc} plots the lines representing the progress of \attr{} vs. test accuracy (complete results in App.~\ref{app:eval-saliency-map-test-acc}). 
SHAP with watermark shows the most consistent and expected increase in \attr{} with test accuracy. For other saliency maps and feature types, the trend is very mild or noisy, suggesting that the attribution method fails to recognize model's increasing reliance on the manipulation in making increasingly accurate predictions. 

\subsection{Attribution vs. Manipulation Visibility}
\textbf{Question: }How well can saliency maps recognize manipulations of different visibility levels? 

\noindent\textbf{Setup: }
We conduct 100 runs, with 20 per manipulation. We further group the 20 runs into 4 groups, with 5 runs in a group using the same manipulation type and effective region but varying degrees of visibility, detailed in App.~\ref{app:eval-saliency-map-visibility}. For example, the visibility for a watermark corresponds to its font size. As before, the labels are reassigned with $r=0.5$ and manipulations applied to the positive class only. 

\noindent\textbf{Expectation: }
A good saliency map should not be affected by manipulation visibility, as long as the model is objectively using it. However, different saliency maps may be better suited to detect more or less visible manipulations. For example, a less visible manipulation may be ignored by the segmentation algorithm used by LIME, while inducing sharper gradients in the decision space. 

\begin{figure}[!htb]
    \centering
    \includegraphics[width=\columnwidth]{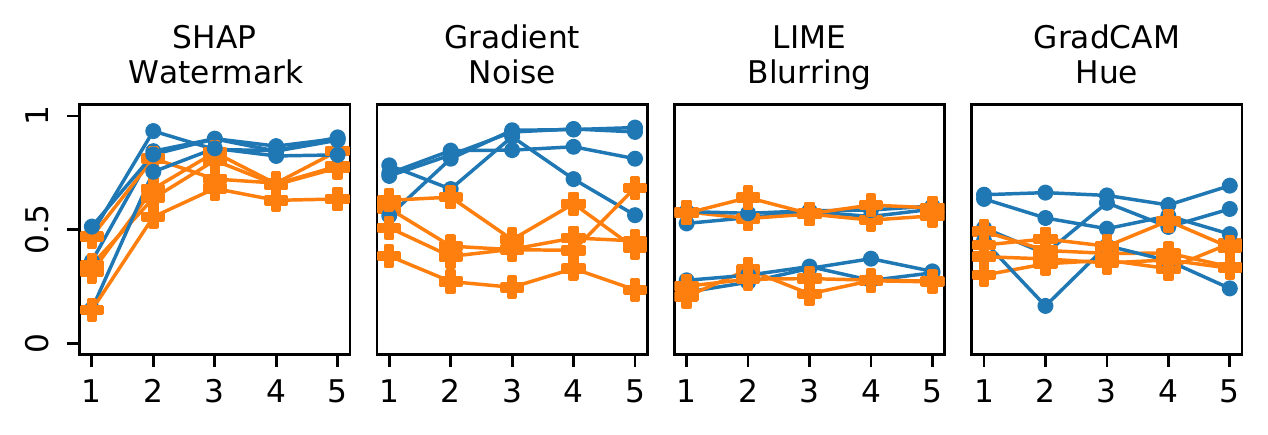}
    \caption{\attr{} vs. feature visibility, more in App.~\ref{app:eval-saliency-map-visibility}. }
    \label{fig:saliency-map-visibility}
\end{figure}

\noindent\textbf{Results: }
Fig.~\ref{fig:saliency-map-visibility} (left) plots each group of five runs as a line, with visibility level on the $x$-axis and \attr{} on the $y$-axis (complete results in App.~\ref{app:eval-saliency-map-visibility}). Except for SHAP on watermark, other methods do not show consistent trend of \attr{} increasing with visibility. While SHAP is more effective on more visible manipulations, we rely most on interpretability methods to uncover precisely the less visible manipulations or artifacts. Unfortunately, none of the methods could satisfy this requirement.

\subsection{Attribution vs. Original Feature Correlation}
\label{sec:saliency-attr-m}

\textbf{Question: }How does the attribution on the manipulation change if the reassigned labels are correlated with the original labels (and thus original input features) to higher or lower degrees (i.e. $r\in [0.5, 1.0]$)? 

\noindent\textbf{Setup: }
For each manipulation, we vary the label reassignment parameter $r\in \allowbreak \{0.5, \allowbreak 0.6, \allowbreak 0.7, \allowbreak 0.8, \allowbreak 0.9, \allowbreak 1.0\}$. For each $r$, we train four models on four class pairs: two of similar species (e.g. class 4 vs. 5 in Fig.~\ref{fig:flickr-bird-overview}) and two of distinct ones (e.g. class 5 vs. 6), for a total of $5\times 6 \times 4 = 120$ runs. 

\noindent\textbf{Expectation: }
For $r > 0.5$, there is no standard definition of the attribution value on the original image features and the manipulations, as any decreasing trend of \attr{} with increasing $r$ is reasonable. However, the Shapley value \citep{roth1988shapley} is a commonly used axiomatic definition for feature attributions. We denote the set of features inside the effective region as $F_M$, for manipulated features, and that outside as $F_O$, for original features. Their Shapley values on performance $v(F_M)$ and $v(F_O)$ are defined as
\begin{align}
    \textstyle v(F_M) &= \textstyle \frac{1}{2} \left[a(F_M){-}a(\varnothing){+}a(F_M\cup F_O){-}a(F_O)\right], \\
    \textstyle v(F_O) &= \textstyle \frac{1}{2} \left[a(F_O){-}a(\varnothing){+}a(F_M\cup F_O){-}a(F_M)\right], 
\end{align}
where $a(F_M), a(F_O), a(\varnothing)$, and $a(F_M\cup F_O)$ refers to the classifier's expected accuracy when only $F_M$, only $F_O$, neither, and both are available, respectively. For a classifier with accuracy $p$, we have $a(\varnothing)=0.5$, $a(F_M)=a(F_M\cup F_O)=p$, and $a(F_O)\leq m$. The formal definition of $a(\cdot)$ and its calculation are in App.~\ref{app:eval-saliency-map-m}. We normalize the Shapley values to $\overline v(F_M)$ and $\overline v(F_O)$ by their sum $\overline v(F_M) + \overline v(F_O)$. 

It is easy to see that
\begin{align}
    \overline v(F_M) &\geq (2p - r - 0.5) / (2p - 1), \label{eq:vfc-ineq}\\
    \overline v(F_O) &\leq (r - 0.5) / (2p - 1). \label{eq:vfo-ineq}
\end{align}
For (near-)perfect classifier with $p\approx 1$, we have $\overline v(F_M)\geq 1.5-r$, and $\overline v(F_O)\leq r - 0.5$. In addition, $a(F_O)$ should be close to $r$ for the distinct pair as the model can better utilize the more distinct original image features, resulting in lower attribution $\overline v(F_M)$ on manipulated features. 

\begin{figure}[!t]
    \centering
    \includegraphics[width=\columnwidth]{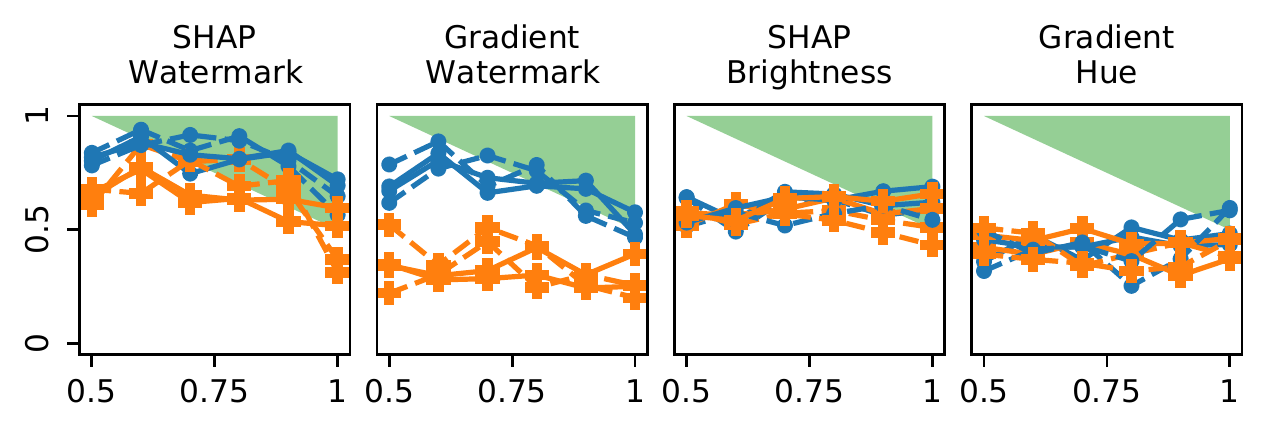}
    \caption{\attr{} vs. $r$, more in App.~\ref{app:eval-saliency-map-m}. Solid/dashed lines represent similar/distinct species pairs. \textcolor[RGB]{44,160,44}{Green} shades represent attribution range per Shapley axioms: \attr{} $\geq 1.5 - r$.}
    \label{fig:saliency-map-m}
\end{figure}

\noindent\textbf{Results: }
All models achieve test accuracy of over 95\%.  Fig.~\ref{fig:saliency-map-m} (right) plots \attr{} vs. $r$ (complete results in App.~\ref{app:eval-saliency-map-m}). Solid lines represent runs with a similar species pair, and dashed lines represent runs with a distinct species pair. The green shaded area represents the area of $\overline v(F_M)\geq 1.5 - r$, the Shapley value range at $p=1$. In other words, values within the green shade are consistent with the Shapley axioms, and those outside are not. Intuitively, for $r$ close to 0.5, the correlation between $F_O$ and the label is very weak, and the (near-)perfect model has to use $F_M$ for high performance, thus \attr{} close to 1. As $r$ increases, the model can choose to rely more heavily on the more-correlating $F_O$ as well, resulting in larger allowable ranges of \attr{}. 

For watermark manipulation, SHAP shows clear decrease in attribution value as $r$ increases, while gradient also tracks the predicted range, but only for the positive class with the manipulation. This trend is not seen in other feature types, even for SHAP which approximates the Shapley values. There does not seem to be a clear difference in attribution values for similar vs. distinct species pairs either. Considering that the set of Shapley \textit{axioms} is commonly accepted as reasonable, it is concerning to see that many saliency maps are inconsistent with it, and important to develop a better understanding about the underlying axiomatic assumptions (if any) made by each of them. 

\subsection{Discussion}
\label{sec:saliency-dis}
Arguably one of the most important application of model explanation is to detect any usage of spurious correlations, but our results cast doubt on this capability from various aspects. We recommend that, before analyzing the actual model, developers should first train models that are guaranteed to use certain known features, and ``dry run'' the planned interpretability methods on them to make sure that these features are indeed highlighted.

\begin{figure*}[!htb]
    \centering
    \includegraphics[width=0.9\textwidth, trim={3.2in 3.4in 0.0in 3.4in}, clip]{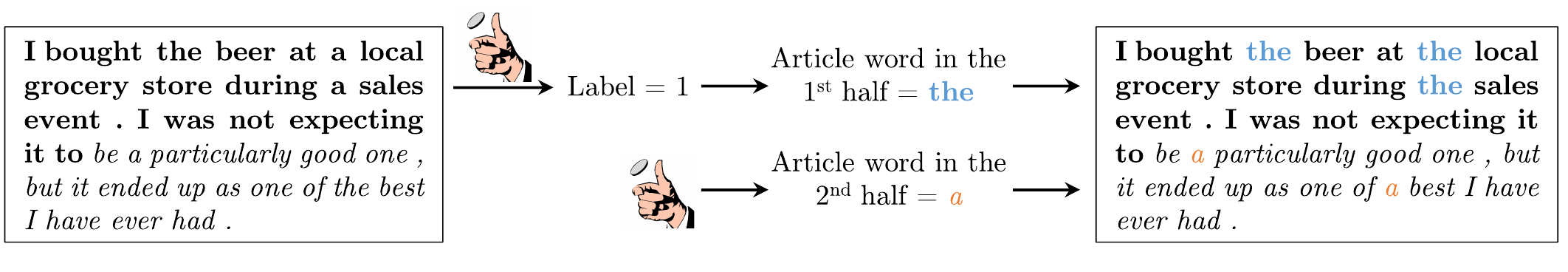}
    \caption{The process to build \textit{CN} dataset for the experiment in Sec.~\ref{sec:rationale-article-mixing}. First, a review is split into two halves at the midpoint, shown in \textbf{bold} and \textit{italics}. Then a label is randomly sampled and assigned to the review. Depending on the label, the articles in the first half are changed to ``a'' or ``the''. They are called \textit{correlating articles}. Then an article word is randomly chosen for the second half, and all articles in the second half are changed to that word. They are called \textit{non-correlating articles}. For \textit{NC} dataset, the roles of two halves are switched. }
    \label{fig:dataset-modification-article-mixing}
\end{figure*}

\section{Evaluating Text Attentions}
\label{sec:eval-attention}

It is known that certain non-semantic features can heavily influence model prediction, such as the email headers \citep{ribeiro2016should}. Plausibly, attention scores should highlight such features, and we rigorously test this with our dataset modification in this section. 

\noindent\textbf{Model: } The attention architecture follows the one used by \citet{wiegreffe2019attention} closely. First, a sentence of $L$ words $(w_1, ..., w_L)$ is converted to a list of 200-dimensional embeddings $(\vec v_1, ..., \vec v_L)$. We use the same embedding data as \citet{lei2016rationalizing} and \citet{bastings2019interpretable}. Then, a Bi-LSTM network builds contextual representations for these words $\vec h_1, ... \vec h_L$, where $\vec h_i\in \mathbb R^{400}$ is the concatenation of the forward and the backward hidden states, each of 200 dimensions. Finally, the attention mechanism computes the sentence representation as
\begin{align}
    \vec k_i &= \mathrm{tanh}(\mathrm{Linear}(\vec h_i)) \in \mathbb R^{200}, \\
    b_i &= \vec q \cdot \vec k_i\\
    a_1, ..., a_L &= \mathrm{softmax}(b_1, ..., b_L), \\
    \vec h &= \sum_{i=1}^L a_i \vec h_i, 
\end{align}
where $\mathrm{Linear}()$ represents a linear layer with learned parameters, $\vec q \in \mathbb R^{200}$ is a learned query vector applied to every sentence, and $a_1, ..., a_L$ are the attention weights for $w_1, ..., w_L$. Finally, a linear layer computes the 2-dimensional logit vector for model prediction. 

\noindent\textbf{Dataset: }We modify the BeerAdvocate dataset \citep{mcauley2012learning} and further select 12,000 reviews split into train, validation, and test sets of sizes 10,000, 1,000 and 1,000 (shuffled differently for each experiment). 

\noindent\textbf{Metric: }The introduced manipulation changes specific words according to the reassigned label. The metric is \attr{} defined on the target words (i.e. effective region). 

\subsection{Highly Obvious Correlating Features}
\label{sec:eval-attention-distinct}

\textbf{Question: }
How well can attention scores focus on highly obvious manipulations? 

\noindent\textbf{Setup: }
From our filtered dataset, we first randomly assign binary labels. For the positive reviews, we change all the article words (\textit{a / an / the}) to ``the'', and for the negative reviews, we change these to ``a''. Thus, only these articles are correlated with the labels and constitute the effective region. 

\noindent\textbf{Expectation: }
Attention of (near-)perfect models should have \attr{} $\approx 1$ to be valid attributions. 

\begin{figure}[!t]
    \centering
    \includegraphics[width=\columnwidth]{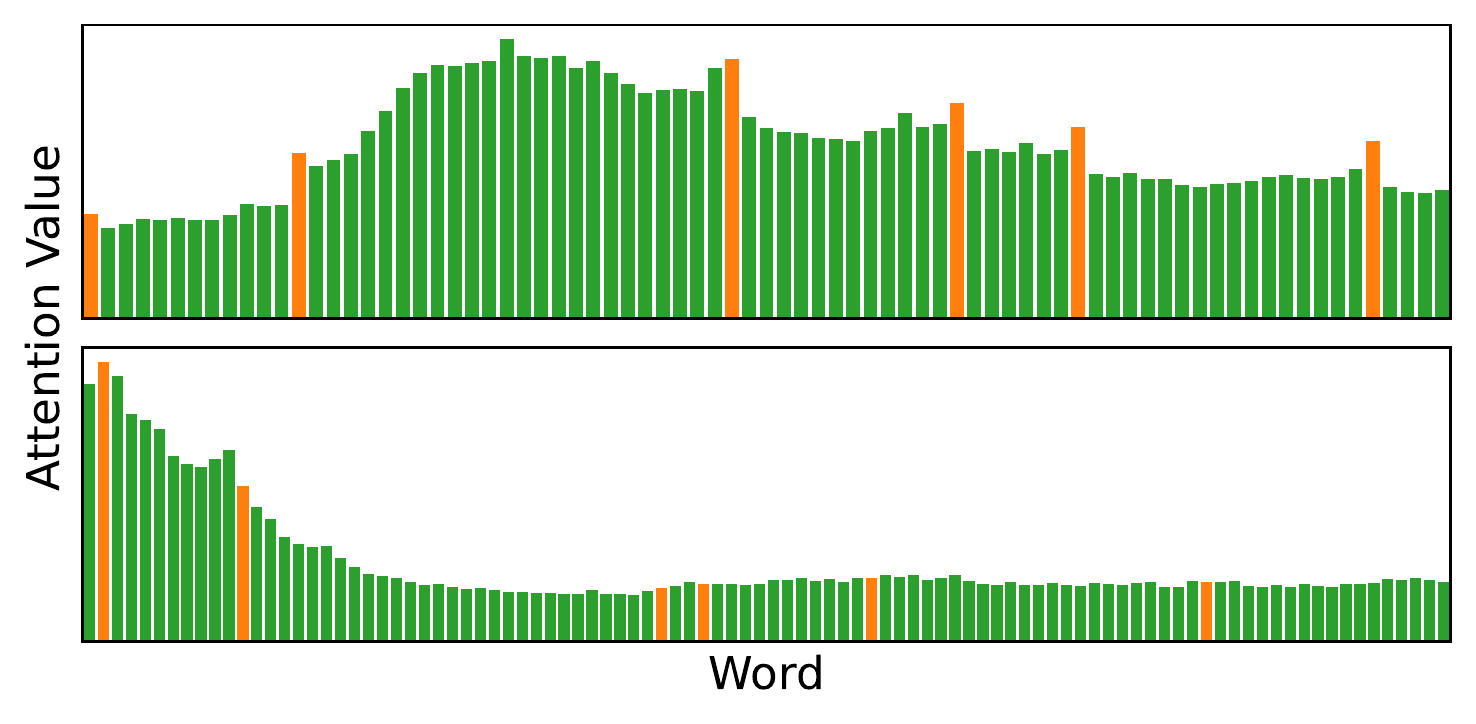}
    \caption{Attention scores for words in two reviews, with more in App.~\ref{app:eval-attention-article}. \textcolor[RGB]{255,127,14}{Orange} and \textcolor[RGB]{44,160,44}{green} bars represent articles and non-articles, respectively. }
    \label{fig:attention-article}
\end{figure}

\noindent\textbf{Results: }
The model achieves over 97\% accuracy. Across the test set, \attr{} on article words is 8.6\%. Considering that articles are 7.9\% of all words, this is better than random, albeit barely. Fig.~\ref{fig:attention-article} visualizes the attention distribution for two reviews, with additional results in Fig.~\ref{fig:attention-article-complete} of App.~\ref{app:eval-attention-article}. Each bars represent weights of words in the review. Green and orange bars represent non-articles and articles respectively. As we can see, the attention on article words either does not stand out from the rest, or at most only locally, \textit{relative} to their neighbors. Generally, there is no strong correlation between high attention values and important words. 

\subsection{Misleading Non-Correlating Features}
\label{sec:eval-attention-mixing}

\textbf{Question: }When some features are known to not correlate with the label but are very similar to correlating ones, do attention scores also focus on these non-correlating ones? 

\noindent\textbf{Setup: }Again from our filtered dataset, we apply two similar manipulations, with only one of them is correlated with the (reassigned) label. Fig.~\ref{fig:dataset-modification-article-mixing} details the construction of two datasets, \textit{CN} and \textit{NC}.

\noindent\textbf{Expectation: }
Same as above. In particular, non-correlating articles should \textit{not} be attended to. 

\begin{figure}[!htb]
\small
    \centering
    \includegraphics[width=\columnwidth]{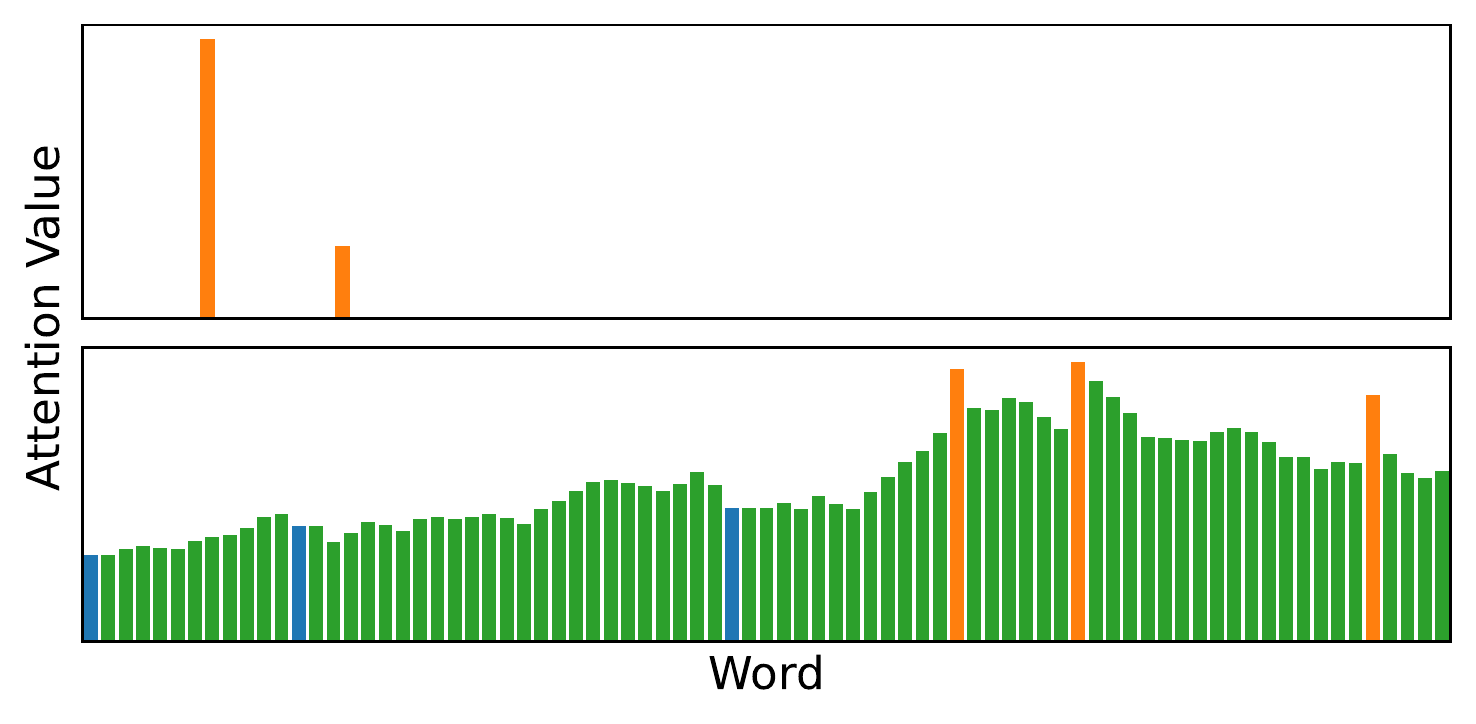}
    \caption{Word attentions for one review in \textit{CN} (top) and \textit{NC} (bottom) dataset, with more in App.~\ref{app:eval-attention-article-mixing}. \textcolor[RGB]{255,127,14}{Orange}, \textcolor[RGB]{31,119,180}{blue}, and \textcolor[RGB]{44,160,44}{green} bars represent correlating articles, non-correlating articles, and other words, respectively. }
    \vspace{-0.2in}
    \label{fig:attention-article-mixing}
\end{figure}

\noindent\textbf{Results: }
The models on both datasets achieve over 97\% accuracy. Fig.~\ref{fig:attention-article-mixing} presents attention visualization, with more in Fig.~\ref{fig:attention-article-mixing-complete} of App.~\ref{app:eval-attention-article-mixing}. The two models show very different behaviors. The \textit{CN} model exclusively focuses attention on correlating articles, while the \textit{NC} model behaves similarly to the previous experiment.

Observing the large variation of behaviors, we further trained the three models ten more times to see if any consistent attention pattern exists. All models achieve over 97\% accuracy. Fig.~\ref{tab:attention-repeat} (left) presents the mean and standard deviation statistics for the 11 runs. The clean attention pattern by the \textit{CN} model does not persist, and the model sometimes assigns higher than random weights on non-correlating articles, especially for the $NC$ dataset. These results further suggests that attention weights cannot be readily and reliably interpreted as attributions without further validation. 

\begin{table*}[tb]
    \centering
    \begin{tabular}{c|c|c|c}\toprule
        Dataset & Corr. Articles & Non-Corr. Articles & Other Words\\\midrule
        Article & $10.3\% \pm \phantom{2}2.4\% \,\,|\,\, 7.9\%$ & NA & $89.7\% \pm \phantom{2}2.4\% \,\,|\,\, 92.1\%$\\
        \textit{CN} & $15.9\% \pm 25.7\% \,\,|\,\, 4.1\%$ & $\phantom{1}5.9\% \pm 4.0\% \,\,|\,\, 3.8\%$ & $78.2\% \pm 24.7\% \,\,|\,\, 92.1\%$\\
        \textit{NC} & $12.0\% \pm \phantom{2}8.4\% \,\,|\,\, 3.8\%$ & $12.6\% \pm 9.3\% \,\,|\,\, 4.1\%$ & $75.4\% \pm 16.7\% \,\,|\,\, 92.1\%$\\\bottomrule
    \end{tabular}
    \caption{Attention attribution statistics across 11 training runs (format: $\mathrm{mean(\%Attr)} \pm \mathrm{stdev(\%Attr)} \mid \mathrm{word\,\,\allowbreak frequency}$). The ``Article'' dataset is the one used in Sec.~\ref{sec:eval-attention-distinct}. }
    \label{tab:attention-repeat}
\end{table*}

\subsection{Discussion}
\label{sec:attention-dis}
Attention is undoubtedly useful as a building block in neural networks, but their \textit{interpretation} as attribution is disputed. Due to the lack of ground truth information on word-prediction correlation, past studies proposed various, and sometimes conflicting, criteria for judging the validity of attribution interpretation \citep{jain2019attention, wiegreffe2019attention, pruthi2020learning}. However, the fundamental correctness of such proxy metrics is unclear. In our studies, we find that attentions can hardly be interpreted as attribution for model understanding and debugging purposes: for most training runs, the attention weights on correlating features at best stand out only \textit{locally}, easily overwhelmed by larger global variations, setting the debate at least on the modified dataset. For natural datasets, we would unavoidably need to rely on proxy metrics, but we recommend future proposals of the metrics to be first calibrated with ground truth in a controlled setting.

\section{Evaluating Text Rationales}
\label{sec:eval-rationale}
In this section, we evaluate rationale models with the same two experiments above (and omit the \textbf{Question} and \textbf{Setup} descriptions). We consider two variants: a reinforcement learning (RL) model \citep{lei2016rationalizing} and a continuous relaxation (CR) model \citep{bastings2019interpretable}. In the original forms, both models regularize the rationale length and continuity. In our experiments, rather than regularizing the length, we train the models to produce rationales that match a target selection rate \sel{}. For a mini-batch of $B$ examples, we use $\lambda \cdot \left|\sum_{i=1}^B \mathrm{len}({\mathrm{rationale}_i)} / \sum_{i=1}^B\mathrm{len}({\mathrm{review}_i)} - \mathrm{Sel}\% \right|$, where $\lambda > 0$ is the regularization strength. Incidentally, we found that the training is much more stable with this regularization, especially for the RL model. We also removed the discontinuity penalty, because ground truth rationales in our experiments are not continuous. 
We use precision and recall metrics as defined in Sec.~\ref{sec:desiderata}.

\subsection{Highly Obvious Manipulations}
\label{sec:rationale-article}

\textbf{Expectation: }
A necessary condition for a non-misleading rationale is that it should include at least one article word, regardless of selection rate. However, a desirable property of rationale is comprehensiveness \citep{yu2019rethinking}: selecting as many article words as possible. Thus, a good rationale model should have high precision when selection rate is low and high recall when selection rate is high. 

\begin{figure}[!htb]
    \centering
    \includegraphics[height=0.5\columnwidth]{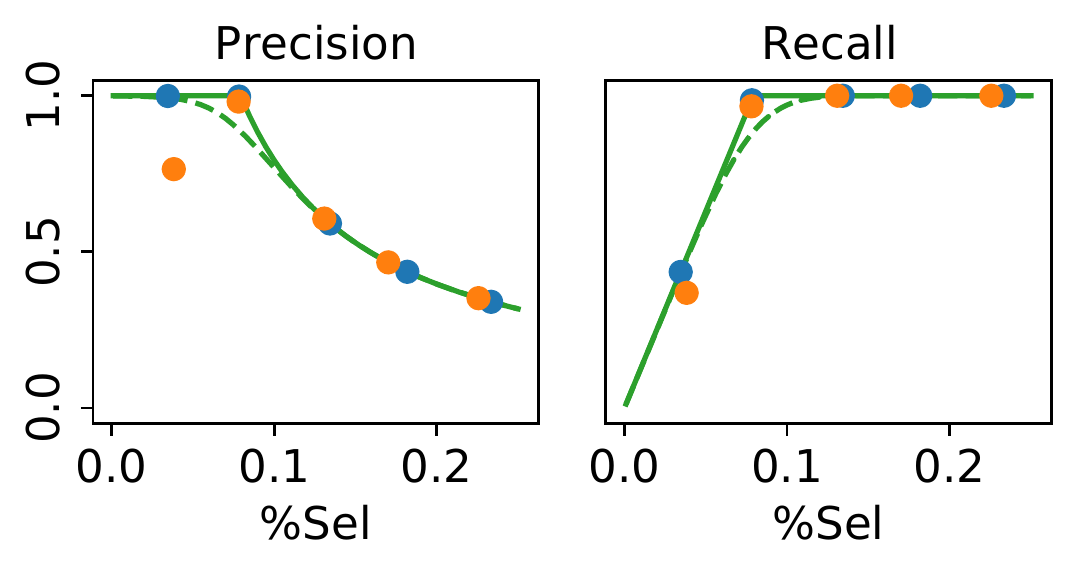}
    
    \begin{tikzpicture}
        \node[draw,text width=0.9\columnwidth] at (0, 0) {\small \textcolor{Orange}{\textbf{\textit{enjoyed}}} @ \textcolor{Orange}{\textbf{\textit{la}}} cave \gb{the} bulles ; simon \& \ri{the} head brewer of brasserie de vines hosted \gb{the} tasting on 11/5 . medium body , frothy mouth-feel , nice carbonation . nice fruity notes upfront , green apples and citrus , with \ri{the} hint of sourness . finishes with \ri{the} fresh piney hop presence and \ri{the} mild bitterness . overall ; great diversity in flavors , very fresh tasting . };
    \end{tikzpicture}

    \caption{\textbf{Top}: Precision and recall for two rationale models in Sec.~\ref{sec:rationale-article}. \textbf{Bottom}: A rationale pattern by the ``faulty'' CR model, selected non-articles in \textcolor{Orange}{\textbf{\textit{orange bold italics}}}, selected articles in \textcolor{Green}{\textbf{green bold}}, and missed articles in \textcolor{Red}{\textit{red italics}}, more in App.~\ref{app:eval-rationale-article}. }
    \label{fig:rationale-article}
\end{figure}

\noindent\textbf{Results: }
We train models with \sel{} $\in\{0.07, \allowbreak 0.09, \allowbreak 0.11, \allowbreak 0.13, 0.15\}$, all with over 97\% accuracy. We evaluate precision and recall of the trained models and plot them in Fig.~\ref{fig:rationale-article} (top) according to the actual rationale selection rate, \sel{}, on the test set. Blue and orange markers are for the RL and CR models respectively. The two green lines show two optimality notions: the solid line enforces aggregate \sel{} for the test set, and the dashed line enforces \sel{} per review. 

Except for the CR model at the lowest \sel{}, all others achieve near-perfect rationale selection on both the precision and and the recall metrics. In particular, they are nearly dataset-wide optimal, due to \sel{} regularization done at the mini-batch level. The ``faulty'' CR model tends to select the first few words consistently, as shown in Fig.~\ref{fig:rationale-article} (bottom) and App.~\ref{app:eval-rationale-article}, but still selects some article words.

\subsection{Misleading Non-Correlating Features}
\label{sec:rationale-article-mixing}

\textbf{Expectation: }Similar to the previous experiment, at least one correlating article word needs to be selected. However, selection of non-correlating articles is arguably more misleading than selection of other non-article words, because it suggests that these non-correlating articles also influence the prediction, even though the classifier simply ignores them. 

\begin{figure}[!htb]
    \centering
    \includegraphics[height=0.5\columnwidth]{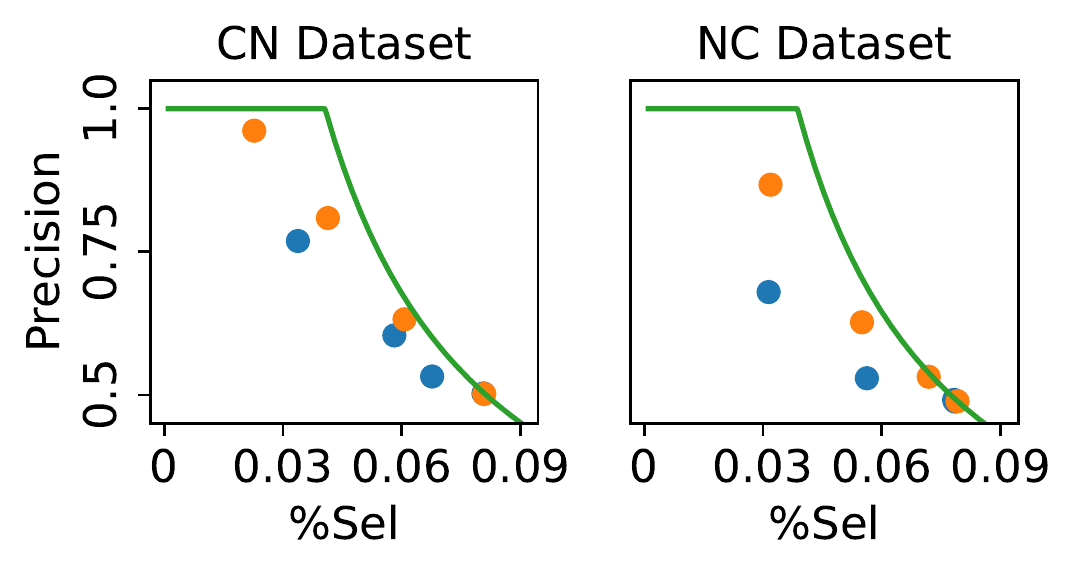}
    \begin{tikzpicture}
        \node[draw,text width=0.9\columnwidth] at (0, 0) {
        \small enjoyed @ la cave \u{\ri{the}} bulles ; simon \& \ri{the} head brewer of brasserie de vines hosted \ri{the} tasting on 11/5 . medium body , frothy mouth-feel , nice carbonation . nice fruity notes upfront , green apples and citrus , with \u{\gb{a}} hint of sourness . finishes with \gb{a} fresh piney hop presence and \u{\gb{a}} mild bitterness . overall ; great diversity in flavors , very fresh tasting . };
    \end{tikzpicture}

    \caption{\textbf{Top}: Precision at different \sel{} for models in Sec.~\ref{sec:rationale-article-mixing}. RL model is in blue and CR in orange. The solid and dashed green lines show optimal metric values when \sel{} is enforced at dataset- and sentence-level. \textbf{Bottom}: A \u{rationale selection} on the \textit{NC} dataset, correlating articles in \textcolor{Green}{\textbf{green bold}} and non-correlating articles in \textcolor{Red}{\textit{red italics}}, more in App.~\ref{app:eval-rationale-article-mixing}. }
    \label{fig:rationale-article-mixing}
\end{figure}

\noindent \textbf{Results: }
We train models with \sel{} $\in\{0.03, \allowbreak 0.05, \allowbreak 0.07, \allowbreak 0.09\}$, all with over 97\% accuracy. Fig.~\ref{fig:rationale-article-mixing} (top) plots the precision of correlating articles for the two datasets, as well as the dataset-wide optimal value. We found the rationales consist of almost exclusively article words. However, especially for the RL model, some correlating articles are missed but non-correlating ones are selected, resulting in markedly less than optimal precision. Fig.~\ref{fig:rationale-article-mixing} (bottom) shows one example for the RL model and additional ones are in App.~\ref{app:eval-rationale-article-mixing}. 

\subsection{Discussion}
\label{sec:rationale-dis}
The structure of rationale models guarantees that causal relationship between the rationale features and the model prediction, but this does not necessarily imply its usefulness to model understanding. Specifically, it could highlight $F_C$ only barely, while including lots of non-correlating $F_N$ (and, in particular, misleading words such as the non-correlating articles)\footnote{There are additional concerns on the unfaithfulness of rationales as Trojan explanations \citep{jacovi2021aligning, zheng2021irrationality}, but they were not identified in our experiments.}. Indeed, our results show that rationale methods are prone to selecting misleading non-correlating features, which obfuscates the model's reasoning process by giving \textit{more} but unnecessary information to the human. The problem is more severe with RL training, possibly due to the known difficulty with REINFORCE \citep{williams1992simple}. Post-processing methods could be developed to further prune rationales to mitigate this problem.

\section{Conclusion and Future Work}

As interpretability methods, especially feature attribution ones, are increasingly deployed for quality assurance of high-stakes systems, it is crucial to ensure these methods work correctly. Current evaluations fall short---primarily due to a lack of clearly defined ground truth. 
Rather than evaluating explanations for models trained on natural datasets, we propose ``unit tests'' to assess whether feature attribution methods are able to uncover ground truth model reasoning on carefully-modified, semi-natural datasets. Surprisingly, none of our evaluated methods across vision and text domains achieve totally satisfactory performance, and we point out various future directions in Sec.~\ref{sec:saliency-dis}, \ref{sec:attention-dis} and \ref{sec:rationale-dis} to improve attribution methods. 

Our dataset modification procedure closely parallels the setup for identifying and debugging model reliance on spurious correlations, which have been known to frequently affect model decisions \citep[e.g.][]{ribeiro2016should, kaushik2019learning, geirhos2020shortcut, jabbour2020deep}. Hence, the mostly negative conclusions cast doubt on this use case of interpretability methods. 

An extension of the proposed evaluation procedure is to move beyond ``artifact'' features, which result from the manual definition of the manipulation function. Given the recent advances on generative modeling such as image inpainting \citep{pathak2016context} and masked language prediction \citep{devlin2019bert}, more realistic features could be generated, perhaps also conditioned on or guided by semantic concepts. This would make the modified dataset much more realistic looking, and thus better simulate another intended use case of interpretability: assisting scientific discovery, in which high-performing models teach humans about features of previous unknown importance.

\section*{Acknowledgement}
An earlier version of this paper was presented at the 2021 NeurIPS Workshop on Explainable AI Approaches for Debugging and Diagnosis. This research is supported by the National Science Foundation (NSF) under the grant IIS-1830282. We thank the reviewers for their reviews. 

\bibliography{references}

\newpage
\appendix

\onecolumn

\section{Relationship to Closely Related Works}
\label{app:rw-comparison}

In this section, we detail the similarity and difference between our proposal and two closely related ones \citep{yang2019benchmarking, adebayo2020debugging}. In terms of similarity, the high-level idea is similar: for a natural dataset, we do not know individual feature contribution, but generally highly correlated and easily discriminative features should have high contribution, and we realize this notion by injecting such features directly. All three works can be seen as operationalizations of this idea.

However, this high-level idea needs two caveats, which set our paper aparts from both works. First, features that we think to be ``easily discriminative'' may not be be considered as such by neural nets. For example, \citet{geirhos2018imagenet} showed that they have particular inclination toward textures. Thus, we can't really trust them to actually pick up and use our introduced features, unless that we are assured that focusing on other features cannot achieve the performance that the model is achieving now. In this aspect, we concretely demonstrated that network trained by \citet{adebayo2018sanity} can achieve good performance when using the ``other features'' exclusively. \citet{yang2019benchmarking} demonstrated this principle, but used out-of-distribution data, so it is not clear whether the failure is due to achieve good performance is really due to the network indeed ignoring the ``other features'' or due to instability of out-of-distribution extrapolation.

The second caveat is that the ``other features'' cannot have any information on the injected features. For example, in Fig. 4 by \citet{adebayo2020debugging}, the saliency map that perfectly crops out the foreground dog shape could imply that the network actually uses the contour of the dog, which is inconsistent with their target conclusion that ``foreground doesn't matter in highly correlated background dataset''. In other words, background with a dog contour cropped out is very informative to the fact that the foreground is a dog. A similar cropping procedure is used by \citet{yang2019benchmarking} as well. Another loophole is discussed in the Remark of Sec.~\ref{sec:modification}, where the lack of one feature at one place could imply the presence of another feature at another place. By comparison, in our work, we define the joint effective region to contain all injected, so we are assured that the features outside of the region absolutely cannot contribute to the high performance, and then evaluate Attr\% for that region.

Finally, our framing is more general. We propose our domain-agnostic framework in Sec.~\ref{sec:desiderata} and \ref{sec:modification}. Thus, it it straightforward to instantiate our idea to other domains such as graph, speech, or time-series data. By comparison, both works are focused on the foreground/background patch setting, and introduces necessary concepts under this context. We also proposed ways to evaluate feature-selection style attributions, more common for text models, at the end of Sec.~\ref{sec:desiderata}.

\section{Additional Details and Results for Saliency Map Evaluations}
\label{app:eval-saliency-map}

\subsection{Manipulation Types}
\label{app:image-correlation}

\begin{figure}[!htb]
    \centering
    \includegraphics[width=0.91\textwidth]{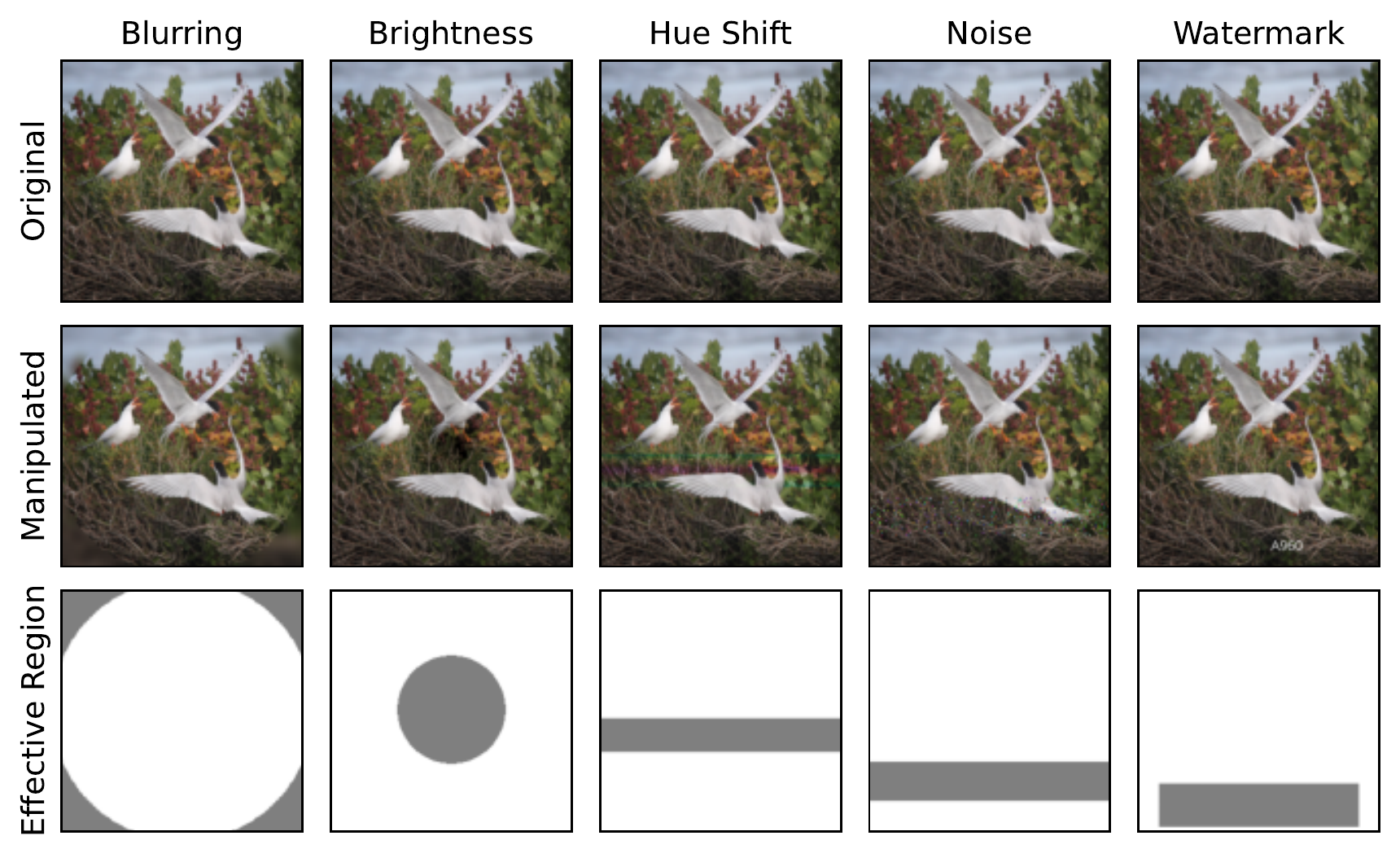}
    \caption{Examples of the five manipulations applied to an image, with respective effective regions (ER) shown in gray. }
    \label{fig:correlation-vis}
\end{figure}

We consider five image manipulation types. These manipulations are designed to simulate possible image artifacts, which an undesirable model may rely on to make decisions. Each manipulation has parameters which define the effective region and the visibility level of the manipulation effect. Some of the manipulation effects are technically stochastic, such as a watermark being placed in a random position, but the effective region captures the localized manipulation effect of all possible random instantiations. The five manipulations are described below, with examples of each manipulation and their associated effective regions shown in Fig.~\ref{fig:correlation-vis}.

\begin{itemize}[leftmargin=0.15in, parsep=3pt]
    \item \textbf{Peripheral blurring} applies a Gaussian filter to the part of the image outside of a certain radius. It is parametrized by
    \begin{itemize}
        \item the radius of the \textit{unaffected} part; and
        \item the standard deviation of the Gaussian blurring filter. 
    \end{itemize}
    Blurring could be due to either camera in motion or artistic post-processing to highlight the main subject of the image.
    \item \textbf{Central brightness shift} gradually changes the brightness in the hue-saturation-brightness (HSB) space inside a certain radius, with maximal change in the center. For our experiments, the brightness change is negative, meaning that the center is dimmed. It is parametrized by
    \begin{itemize}
        \item the radius of the dimmed region; and
        \item the magnitude of the brightness shift at the center. 
    \end{itemize}
    Brightness shift could be due to times of the day, or the use of artificial light to illuminate the subject.
    \item \textbf{Striped hue shift} modifies the hue (i.e. color) value of a vertical stripe in the image. From top to bottom in the stripe, the hue value is first increased and then decreased in a sinusoidal pattern. It is parametrized by
    \begin{itemize}
        \item the upper position of the stripe; 
        \item the lower position of the stripe, with the width of the stripe being (upper - lower); and
        \item the magnitude of sinusoidal pattern. 
    \end{itemize}
    Hue shift could be due to errors in conversion of different color space encodings, which may result in color loss or distortion.
    \item \textbf{Striped noise} randomly changes pixels inside a vertical stripe to a uniformly random RGB value. It is parametrized by
    \begin{itemize}
        \item the upper position of the stripe; 
        \item the lower position of the stripe, with the width of the stripe being (upper - lower); and
        \item the probability that each pixel is replaced. 
    \end{itemize}
    Pixel noise could be due to lossy compression or data loss during transmission. 
    \item \textbf{Watermark} overlays a text reading ``IMGxxxx'', where ``xxxx'' are four random digits, to a random location inside a rectangular region. ``IMG'' is written in white and the digits are written in black. It is parametrized by
    \begin{itemize}
        \item the upper-left coordinate of the rectangular region; 
        \item the lower-right coordinate of the rectangular region; and
        \item the font size of the watermark text. 
    \end{itemize}
    Watermark is a commonly employed technique to attribute the author/organization of the image. 
\end{itemize}

\noindent Normally, none of them should be expected to correlate with the label. However, especially with image scraping on the web and crowdsourced dataset construction, it is possible that some spurious correlations leak into the final dataset.

\subsection{Saliency Map Methods}
\label{app:eval-saliency-map-methods}

We consider five saliency map methods. 
\begin{itemize}[leftmargin=10pt, parsep=3pt]
    \item \textbf{Gradient} \citep{simonyan2013deep} computes the gradient of the logit for the predicted class with respect to the input image. The three channels of gradient are summed up in absolute value to get a single channel. 
    \item \textbf{SmoothGrad} \citep{smilkov2017smoothgrad} averages the gradients on 50 copies of the input image $I$, each injected with independent Gaussian noise with with $\mu=0$ and $\sigma=0.15\cdot (\max I - \min I)$, where $\max I$ and $\min I$ are the maximal and minimal pixel values of the image. 
    \item \textbf{GradCAM} \citep{selvaraju2017grad} computes a saliency map from convolution filter responses. Since we use the fully convolutional ResNet-34, this method reduces to the class activation mapping (CAM) \citep{zhou2016learning}. 
    \item \textbf{LIME} \citep{ribeiro2016should} performs a linear regression using super-pixels of the input image. The absolute values of the coefficients are used to derive the saliency map. We use the default implementation of \texttt{lime.lime\_} \texttt{image.LimeImageExplainer} with the quickshift clustering as the super-pixel segmentation algorithm. 
    \item \textbf{SHAP} \citep{lundberg2017unified} uses the idea of Shapley value \citep{roth1988shapley} for attribution. We use the GradientSHAP instantiation with the default setting of \texttt{shap.GradientExplainer}. We us the entire test set as the ``background'' data. 
\end{itemize}

\newpage

\subsection{Attribution vs. Effective Region Size}
\label{app:eval-saliency-map-attr-ba}
\begin{figure}[!htb]
    \centering
    \includegraphics[width=\textwidth]{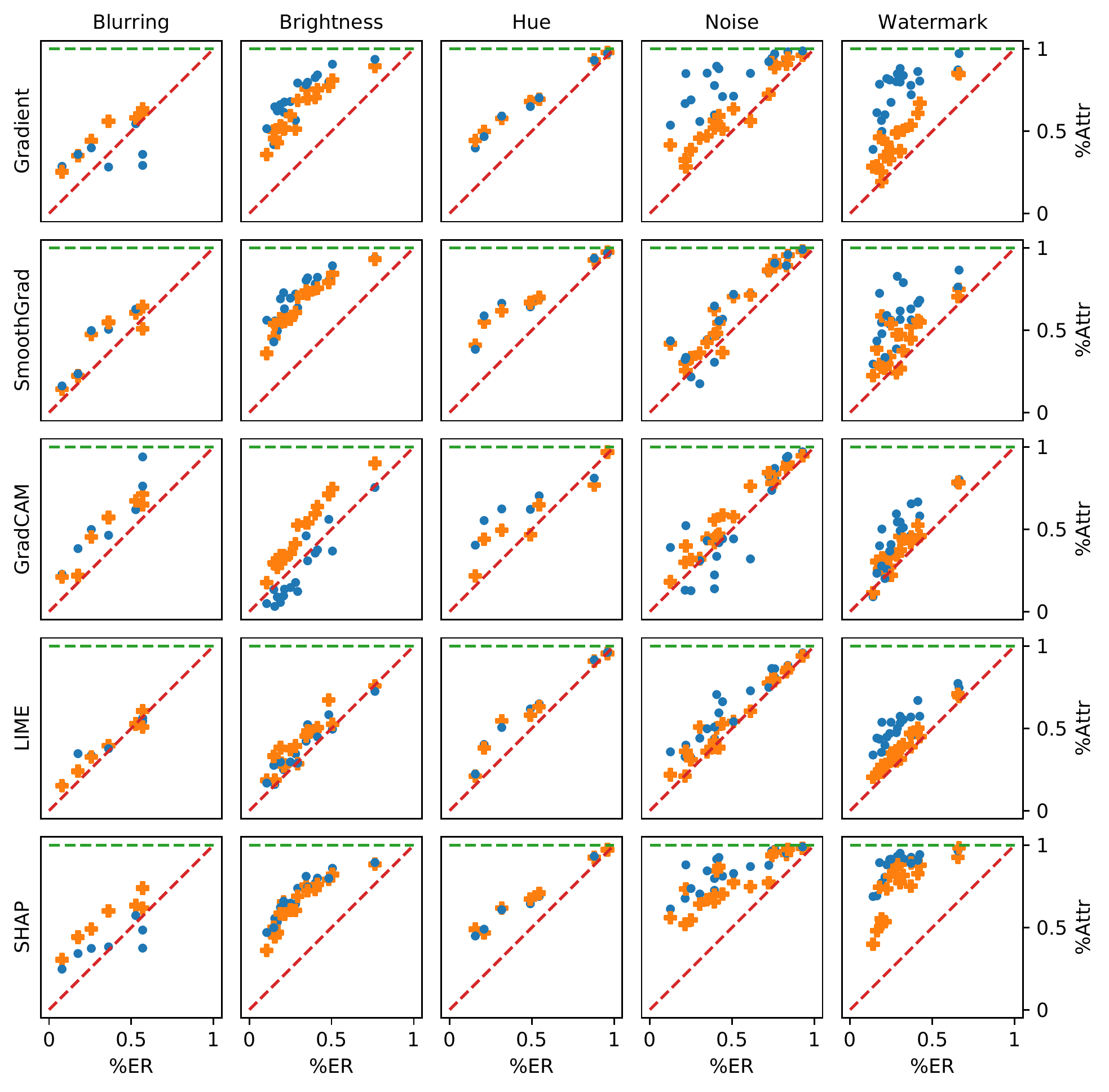}
    \caption{\attr{} vs \er{} for all pairs of saliency maps and manipulations. }
    \label{fig:saliency-map-attr-ba-complete}
\end{figure}

Fig.~\ref{fig:saliency-map-attr-ba-complete} shows \attr{} vs \er{} for all pairs of saliency maps and manipulations.

\newpage
\FloatBarrier

\subsection{Attribution vs. Test Accuracy}
\label{app:eval-saliency-map-test-acc}
\begin{figure}[!htb]
    \centering
    \includegraphics[width=\textwidth]{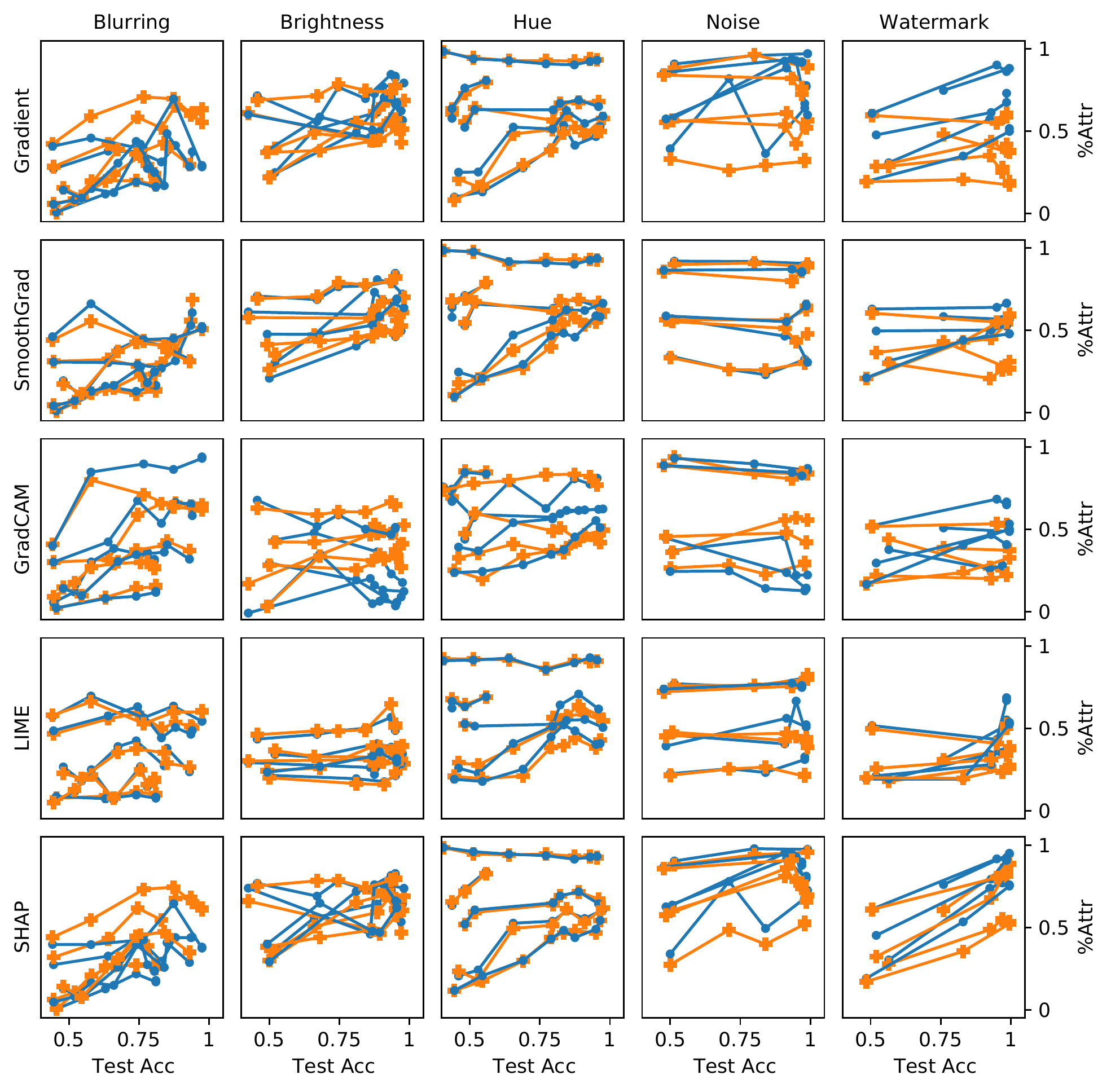}
    \caption{\attr{} vs test accuracy for all pairs of saliency maps and manipulations. }
    \label{fig:saliency-map-test-acc-complete}
\end{figure}

Fig.~\ref{fig:saliency-map-test-acc-complete} shows \attr{} vs test accuracy for all pairs of saliency maps and manipulations. 

\newpage
\FloatBarrier

\subsection{Attribution vs. Manipulation Visibility}
\label{app:eval-saliency-map-visibility}
We define the five visibility levels for each manipulation type as below. Note that for fair comparison of \attr{} at different visibility levels, it is crucial that the effective regions are independent of the visibility, which is satisfied in all manipulation types below. 
\begin{itemize}[leftmargin=10pt, parsep=3pt]
    \item \textbf{Blurring: }The visibility level is defined as the Gaussian blur standard deviation, with values of $\{2, 4, 6, 8, 10\}$ pixels, from least visible to most. 
    \item \textbf{Brightness: }The visibility level is defined as the magnitude of the brightness shift, with values of $\{0.1, \allowbreak 0.15, \allowbreak 0.2, \allowbreak 0.25, \allowbreak 0.3\}$ brightness component of the color (in the range of [0, 1]), from least visible to most. 
    \item \textbf{Hue: }The visibility is defined as the magnitude of the hue shift, with values of $\{0.05, \allowbreak 0.1, \allowbreak 0.15, \allowbreak 0.2, \allowbreak 0.25\}$ hue component of the color (in the range of [0, 1]), from least visible to most. 
    \item \textbf{Noise: }The visibility is defined as the probability that a pixel is replaced by a random value, with values of $\{0.02, \allowbreak 0.04, \allowbreak 0.06, \allowbreak 0.08, \allowbreak 0.1\}$, from least visible to most. 
    \item \textbf{Watermark: }The visibility is defined as the font size of the watermark, with values of $\{7, 9, 11, 13, 15\}$ pixels, from least visible to most. 
\end{itemize}

\begin{figure}[!htb]
    \centering
    \includegraphics[width=\textwidth]{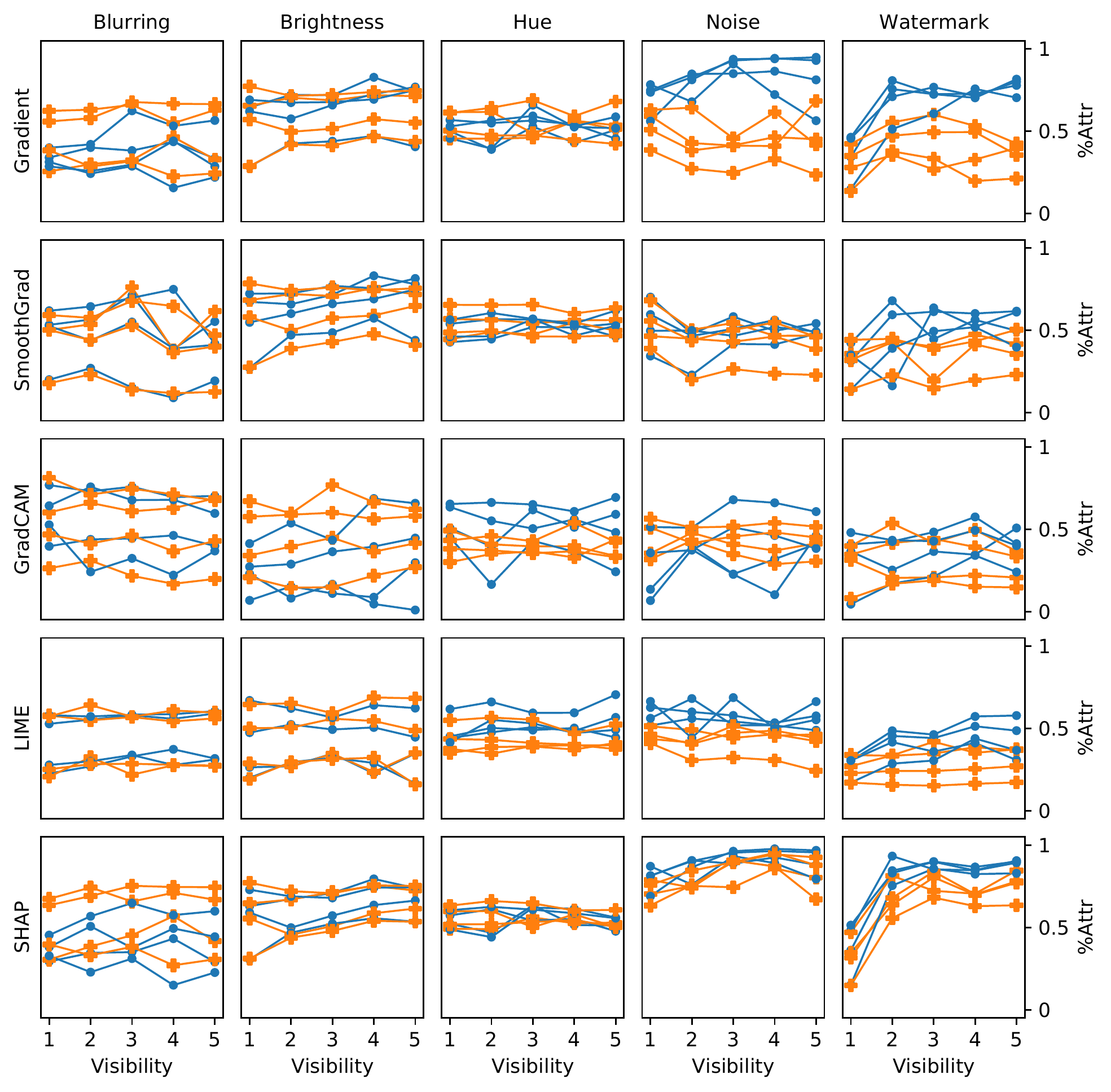}
    \caption{\attr{} vs manipulation visibility for all pairs of saliency maps and manipulations. }
    \label{fig:saliency-map-visibility-complete}
\end{figure}

Fig.~\ref{fig:saliency-map-visibility-complete} shows \attr{} vs manipulation visibility for all pairs of saliency maps and manipulations. 

\subsection{Attribution vs. Original Feature Correlation}
\label{app:eval-saliency-map-m}

As explained in Sec.~\ref{sec:saliency-attr-m}, $a(F)$ represents the expected accuracy of the model when given only feature $F$. Note that this value should \textit{not} be calculated as the model accuracy on images with every pixel but $F$ being blacked out, because such images are out of distribution where the model may exhibit unreasonable behaviors (c.f. discussion by \citet{hooker2018benchmark}). 

Instead, the suppression of information beyond $F$ can be understood as an inability for the model to distinguish inputs that agree on $F$. This leads to the following process of simulating such a prediction. First, let $\mathbb P_{X, Y|F=f}$ be the data distribution conditioned on $F=f$. Since all features other than $F$ are suppressed, the model cannot further distinguish two inputs $x, x'\sim \mathbb P_{X|F=f}$. As a result, the expected accuracy can be computed by comparing the model's prediction on $x$ against the ground truth label on $x'$. Then we take the expectation of this accuracy according to different values of $f\sim \mathbb P_F$, where $\mathbb P_F$ is the marginal distribution of $F$. Formally, for the model prediction function $g: \mathcal X\rightarrow \mathcal Y$, we have
\begin{align}
    a(F) = \mathbb E_{f\sim \mathbb P_F}\left[\mathbb E_{x, y\sim \mathbb P_{X, Y|F=f}}\left[\mathbb E_{x', y'\sim \mathcal P_{X,  Y | F=f}} \left[\mathbbm 1_{g(x)=y'}\right]\right]\right]. 
\end{align}

With balanced label distribution, $a(\varnothing)$ means that the model has no information about the input, and thus the accuracy is 0.5. On the other hand, $a(F_M\cup F_O)$ means that the model has full access to the input, and thus the accuracy is the normal model accuracy $p$. In addition, we have $a(F_O)\leq r$, because the label reassignment weakens the correlation between $F_O$ and the label. 

Finally and somewhat counter-intuitively, the above definition also implies that $a(F_M)=a(F_M\cup F_O)=p$ for the following reason: since every $F_M=f_M$ is perfectly correlated with the label, all the data in $\mathbb P_{X, Y|F_M=f_M}$ have the same label and thus the non-identifiability of any two inputs $x$ and $x'$ does not additionally degrade the model performance. However, note that this result comes from the mechanical application of Shapley value calculation, which is a popular and \textit{axiomatic} definition of attribution. Whether it is reasonable in light of this implication is beyond the scope of the paper. 

\begin{figure}[!htb]
    \centering
    \includegraphics[width=\textwidth]{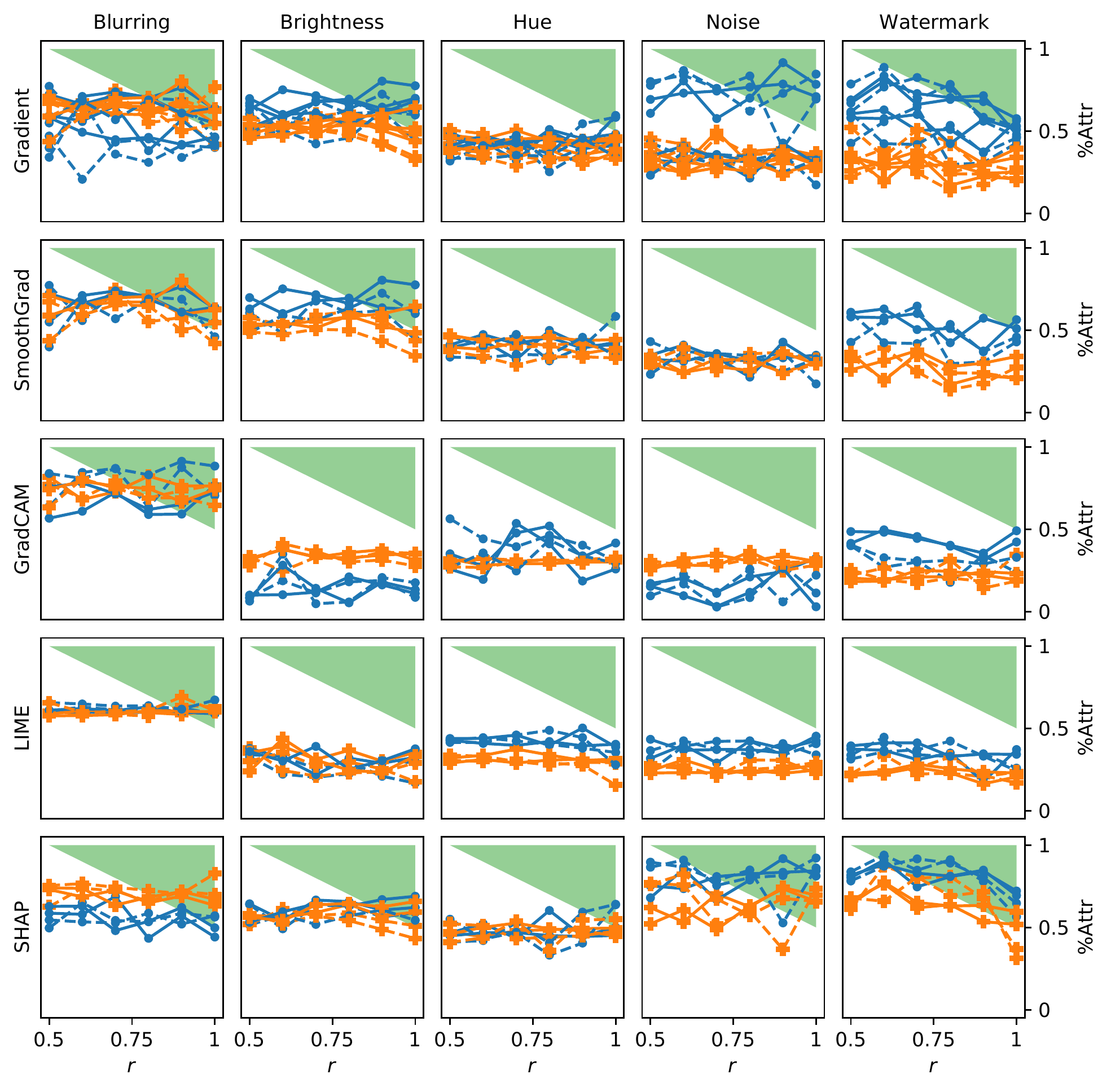}
    \caption{\attr{} vs label reassignment parameter $r$ for all pairs of saliency maps and manipulations. }
    \label{fig:saliency-map-m-complete}
\end{figure}

Fig.~\ref{fig:saliency-map-m-complete} shows \attr{} vs the label reassignment parameter $r$ for all pairs of saliency maps and manipulations. 

\FloatBarrier

\clearpage

\section{Additional Results for Attention Mechanism Evaluations}
\label{app:eval-attention}

\subsection{Highly Obvious Manipulations}
\label{app:eval-attention-article}
\begin{figure}[!htb]
    \centering
    \includegraphics[width=0.95\columnwidth]{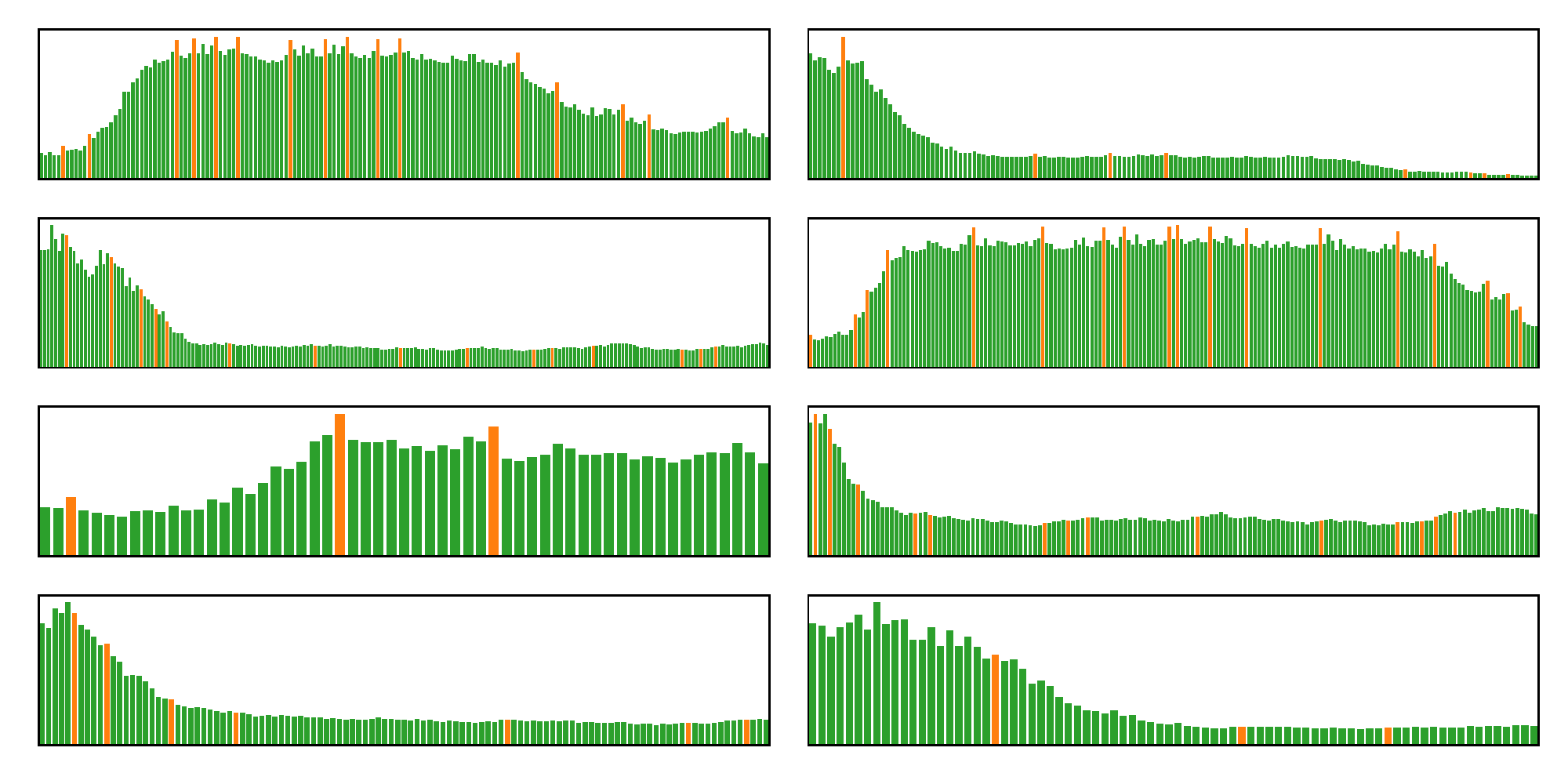}
    \caption{Additional attention distributions. Orange/green bars represent articles/non-articles. }
    \label{fig:attention-article-complete}
\end{figure}

Fig.~\ref{fig:attention-article-complete} presents additional visualizations of the learned attention distribution of the model. 

\subsection{Misleading Non-Correlating Features}
\label{app:eval-attention-article-mixing}
\begin{figure}[!htb]
    \centering
    \includegraphics[width=0.95\columnwidth]{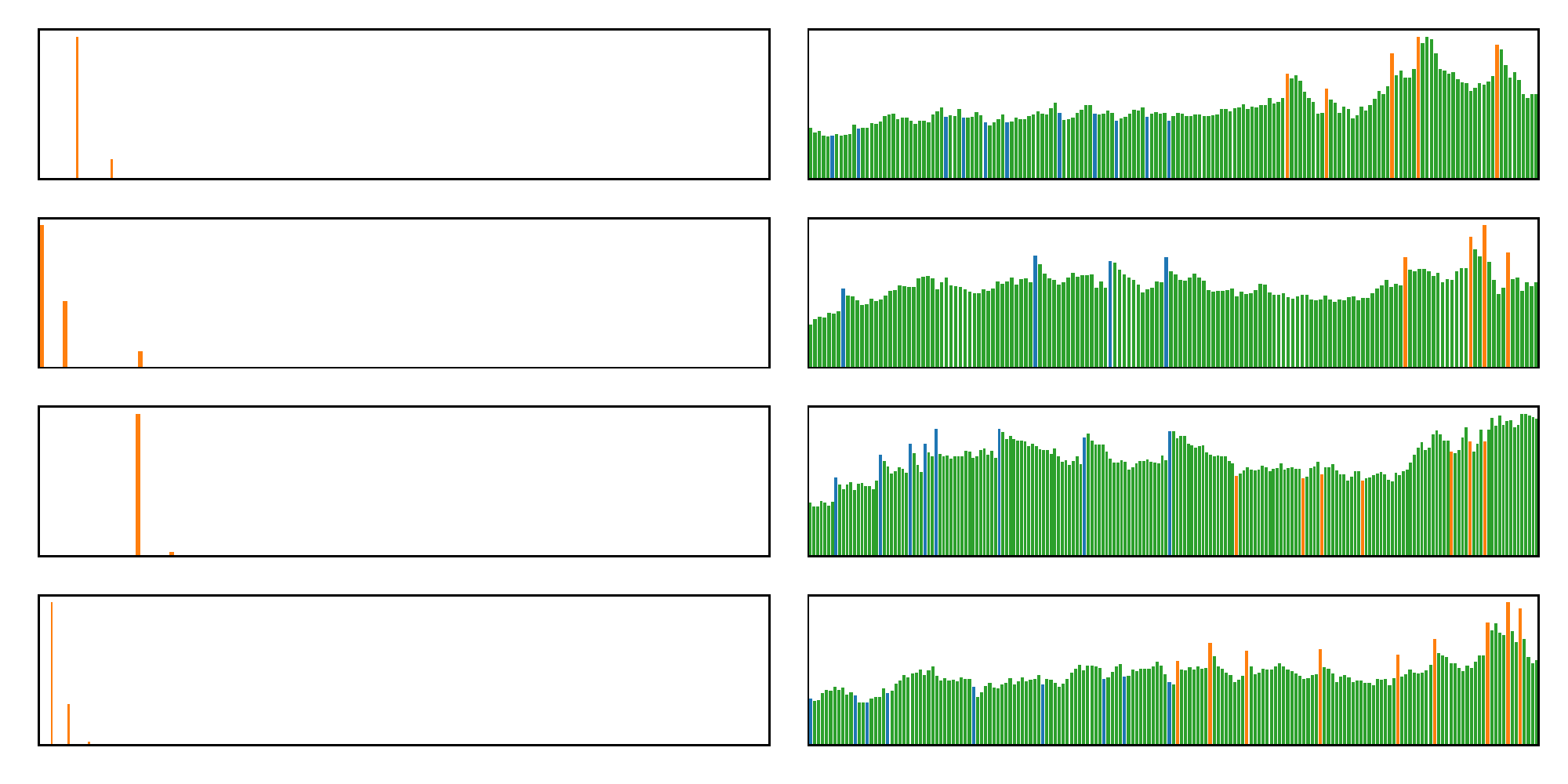}
    \caption{Additional attention visualizations on the \textit{CN} (left) and \textit{NC} (right) datasets. Orange/blue/green bars represent corr. articles/ non-corr. articles/other words. }
    \label{fig:attention-article-mixing-complete}
\end{figure}

Fig.~\ref{fig:attention-article-mixing-complete} presents additional visualizations of the learned attention distribution on the \textit{CN} (left) and \textit{NC} (right) datasets. 

\newpage
\section{Additional Results for Rationale Model Evaluations}
\label{app:eval-rationale}

\subsection{Highly Obvious Manipulations}
\label{app:eval-rationale-article}
Fig.~\ref{fig:rationale-article-complete} presents four additional reviews annotated by the ``faulty'' CR model showing that it consistently selects the first few words of the review. 
\begin{figure}[!htb]
    \centering
    \begin{tabular}{p{3in}|p{3in}}\toprule
        \textcolor{Orange}{\textbf{\textit{pours}}} \textcolor{Green}{\textbf{a}} clear yellow . 1/4 inch head of \textcolor{Green}{\textbf{a}} white color . slight retention and slight lacing . smells of sweet malt , pale malt , fruit , and slight bread aroma . fits 
        \textcolor{Red}{\textit{a}} style of \textcolor{Green}{\textbf{a}} belgian pale ale . mouth feel is smooth and crisp with \textcolor{Red}{\textit{a}} high carbonation level . tastes of pale malt , yeast cleanliness , slight hops , and very slight fruit . overall , \textcolor{Red}{\textit{a}} decent brew but nothing special . 
 & \textcolor{Orange}{\textbf{\textit{bottle to}}} snifter glass . pitch black with little lacing around edge . smells like \textcolor{Red}{\textit{the}} typical oatmeal stout . taste has \textcolor{Red}{\textit{the}} great balance between both milk and oatmeal . sweet from \textcolor{Red}{\textit{the}} sugars and mild dark chocolate in \textcolor{Red}{\textit{the}} after taste . smooth and chewey . leans to \textcolor{Red}{\textit{the}} heavier side in \textcolor{Red}{\textit{the}} mouth . great example of two styles blended . worth seeking . 
 \\\midrule
       \textcolor{Orange}{\textbf{\textit{12oz can}}} poured \textcolor{Orange}{\textbf{\textit{into}}} pint glass . pours \textcolor{Green}{\textbf{the}} pale golden straw color with \textcolor{Green}{\textbf{the}} 2 finger fizzy head that settles quickly . slightly hazy when held to \textcolor{Green}{\textbf{the}} light . smell is fairly nuetral with \textcolor{Red}{\textit{the}} bit of sweet malts coming through . \textcolor{Red}{\textit{the}} slight scent of something metallic . taste is decent . nothing crazy or unique but extremely clean , classic american pale lager flavor . mild light malt flavor with just enough hops for balance . no major off-flavors here . mouthfeel is fluid and crisp . this went down quickly and i am not \textcolor{Green}{\textbf{the}} fizzy yellow beer fan . for what it is , it 's done well .   
       & \textcolor{Orange}{\textit{\textbf{a- dark}}} brown with hints of amber at \textcolor{Green}{\textbf{a}} edges , small head which disappeared quickly and dissipated into \textcolor{Green}{\textbf{a}} few sad bubbles . s- tons of sweet bourbon booze . raisins , sugared malts , dark chocolate filled with raspberry . t- booze and brown sugared malts mingle with one another . this is drinking like \textcolor{Red}{\textit{a}} barleywine to me . lots of wood and oak flavor drenched in booze . m- smooth creamy with enough carbonation . d- it 's \textcolor{Red}{\textit{a}} delicious brew that needs to be savored one of \textcolor{Green}{\textbf{a}} best if not \textcolor{Green}{\textbf{a}} best scotch ales ive had . \\\bottomrule
    \end{tabular}
    \caption{4 additional reviews annotated by the ``faulty'' continuous relaxation model that consistently selects the first few words  regardless. Selected non-articles in \textcolor{Orange}{\textbf{\textit{orange bold italics}}}, selected articles in \textcolor{Green}{\textbf{green bold}}, and missed articles in \textcolor{Red}{\textit{red italics}}. }
    \label{fig:rationale-article-complete}
\end{figure}

\subsection{Misleading Non-Correlating Features}
\label{app:eval-rationale-article-mixing}
Fig.~\ref{fig:rationale-article-mixing-complete} shows rationale selections by the two models for the same review at the same target \sel{}. 
\begin{figure}[!htb]
    \centering
    \begin{tabular}{c|p{3in}|p{3in}}\toprule
    & \multicolumn{1}{c|}{\textit{CN} Dataset} & \multicolumn{1}{c}{\textit{NC} Dataset}\\\midrule
    \multirow{7}{*}{\rotatebox[origin=c]{90}{CR Model}}    & enjoyed @ la cave \u{\gb{a}} bulles ; simon \& \u{\gb{a}} head brewer of brasserie de vines hosted \u{\gb{a}} tasting on 11/5 . medium body , frothy mouth-feel , nice carbonation . nice fruity notes upfront , green apples and citrus , with \ri{the} hint of sourness . finishes with \ri{the} fresh piney hop presence and \ri{the} mild bitterness . overall ; great diversity in flavors , very fresh tasting . 
    & enjoyed @ la cave \ri{the} bulles ; simon \& \ri{the} head brewer of brasserie de vines hosted \ri{the} tasting on 11/5 . medium body , frothy mouth-feel , nice carbonation . nice fruity notes upfront , green apples and citrus , with \gb{a} hint of sourness . finishes with \u{\gb{a}} fresh piney hop presence and \u{\gb{a}} mild bitterness . overall ; great diversity in flavors , very fresh tasting . 
        \\\midrule
    \multirow{7}{*}{\rotatebox[origin=c]{90}{RL Model}}  & enjoyed @ la cave \u{\gb{a}} bulles ; simon \& \u{\gb{a}} head brewer of brasserie de vines hosted \u{\gb{a}} tasting on 11/5 . medium body , frothy mouth-feel , nice carbonation . nice fruity notes upfront , green apples and citrus , with \u{\ri{the}} hint of sourness . finishes with \u{\ri{the}} fresh piney hop presence and \u{\ri{the}} mild bitterness . overall ; great diversity in flavors , very fresh tasting .  
    & enjoyed @ la cave \u{\ri{the}} bulles ; simon \& \ri{the} head brewer of brasserie de vines hosted \ri{the} tasting on 11/5 . medium body , frothy mouth-feel , nice carbonation . nice fruity notes upfront , green apples and citrus , with \u{\gb{a}} hint of sourness . finishes with \gb{a} fresh piney hop presence and \u{\gb{a}} mild bitterness . overall ; great diversity in flavors , very fresh tasting . \\\bottomrule
    \end{tabular}
    \caption{Additional reviews from \textit{CN} and \textit{NC} datasets for the two models. Selected words are \underline{underlined}. Ground truth correlating articles are in \textcolor{Green}{\textbf{green bold}}, and non-correlating articles in \textcolor{Red}{\textit{red italics}}. The CR model performs well on this review, focusing exclusively on correlating articles, while the RL model selects non-correlating articles, and misses a correlating one for the \textit{NC} dataset. }
    \label{fig:rationale-article-mixing-complete}
\end{figure}

\end{document}